\documentclass[sigconf,natbib=true,anonymous=false]{acmart}

\usepackage{multirow}
\usepackage{threeparttable}
\usepackage[ruled]{algorithm2e}
\usepackage{caption}
\usepackage{subcaption}
\usepackage{makecell}
\newcommand{\modelname}{MICQA}
\newcommand{\rerankname}{post-ranker}
\AtBeginDocument{%
  \providecommand\BibTeX{{%
    \normalfont B\kern-0.5em{\scshape i\kern-0.25em b}\kern-0.8em\TeX}}}

\setcopyright{acmcopyright}
\copyrightyear{2018}
\acmYear{2018}
\acmDOI{10.1145/1122445.1122456}

\acmConference[Woodstock '18]{Woodstock '18: ACM Symposium on Neural
  Gaze Detection}{June 03--05, 2018}{Woodstock, NY}
\acmBooktitle{Woodstock '18: ACM Symposium on Neural Gaze Detection,
  June 03--05, 2018, Woodstock, NY}
\acmPrice{15.00}
\acmISBN{978-1-4503-XXXX-X/18/06}

\begin{document}

\title{Multifaceted Improvements for Conversational Open-Domain Question Answering}
\settopmatter{authorsperrow=3}


\author{Tingting Liang}
\authornote{Both authors contributed equally to this research.}
\affiliation{%
  \institution{Hangzhou Dianzi University}
  \city{Hangzhou}
  \country{China}
}
\email{liangtt@hdu.edu.cn}

\author{Yixuan Jiang}
\authornotemark[1]
\affiliation{%
  \institution{Hangzhou Dianzi University}
  \city{Hangzhou}
  \country{China}
}
\email{jyx201050027@hdu.edu.cn}

\author{Congying Xia}
\affiliation{%
  \institution{University of Illinois at Chicago}
  \city{Chicago}
  \country{US}}
\email{cxia8@uic.edu}

\author{Ziqiang Zhao}
\affiliation{%
  \institution{Hangzhou Dianzi University}
  \city{Hangzhou}
  \country{China}
}
\email{zhaoziqiang@hdu.edu.cn}

\author{Yuyu Yin}
\authornote{Corresponding author}
\affiliation{%
  \institution{Hangzhou Dianzi University}
  \city{Hangzhou}
  \country{China}
}
\email{yinyuyu@hdu.edu.cn}

\author{Philip S. Yu}
\affiliation{%
  \institution{University of Illinois at Chicago}
  \city{Chicago}
  \country{US}}
\email{psyu@uic.edu}

\renewcommand{\shortauthors}{Liang and Jiang, et al.}

\begin{abstract}
Open-domain question answering (OpenQA) is an important branch of textual QA which discovers answers for the given questions based on a large number of unstructured documents. Effectively mining correct answers from the open-domain sources still has a fair way to go. Existing OpenQA systems might suffer from the issues of question complexity and ambiguity, as well as insufficient background knowledge.
Recently, conversational OpenQA is proposed to address these issues with the abundant contextual information in the conversation.
Promising as it might be, there exist several fundamental limitations including the inaccurate question understanding, the coarse ranking for passage selection, and the inconsistent usage of golden passage in the training and inference phases.
To alleviate these limitations, in this paper, we propose a framework with Multifaceted Improvements for Conversational open-domain Question Answering ({\modelname}). Specifically, {\modelname} has three significant advantages. First, the proposed KL-divergence based regularization is able to lead to a better question understanding for retrieval and answer reading. Second, the added post-ranker module can push more relevant passages to the top placements and be selected for reader with a two-aspect constrains. Third, the well designed curriculum learning strategy effectively narrows the gap between the golden passage settings of training and inference, and encourages the reader to find true answer without the golden passage assistance.
Extensive experiments conducted on the publicly available dataset OR-QuAC demonstrate the superiority of {\modelname} over the state-of-the-art model in conversational OpenQA task.


\end{abstract}

\ccsdesc[500]{Computer systems organization~Embedded systems}
\ccsdesc[300]{Computer systems organization~Redundancy}
\ccsdesc{Computer systems organization~Robotics}
\ccsdesc[100]{Networks~Network reliability}

\keywords{Open-Domain, Conversational Question Answering, Curriculum Learning}

\maketitle

\section{Introduction}
\begin{figure*}
     \centering
     \begin{subfigure}[b]{0.33\textwidth}
         \centering
         \includegraphics[width=\textwidth]{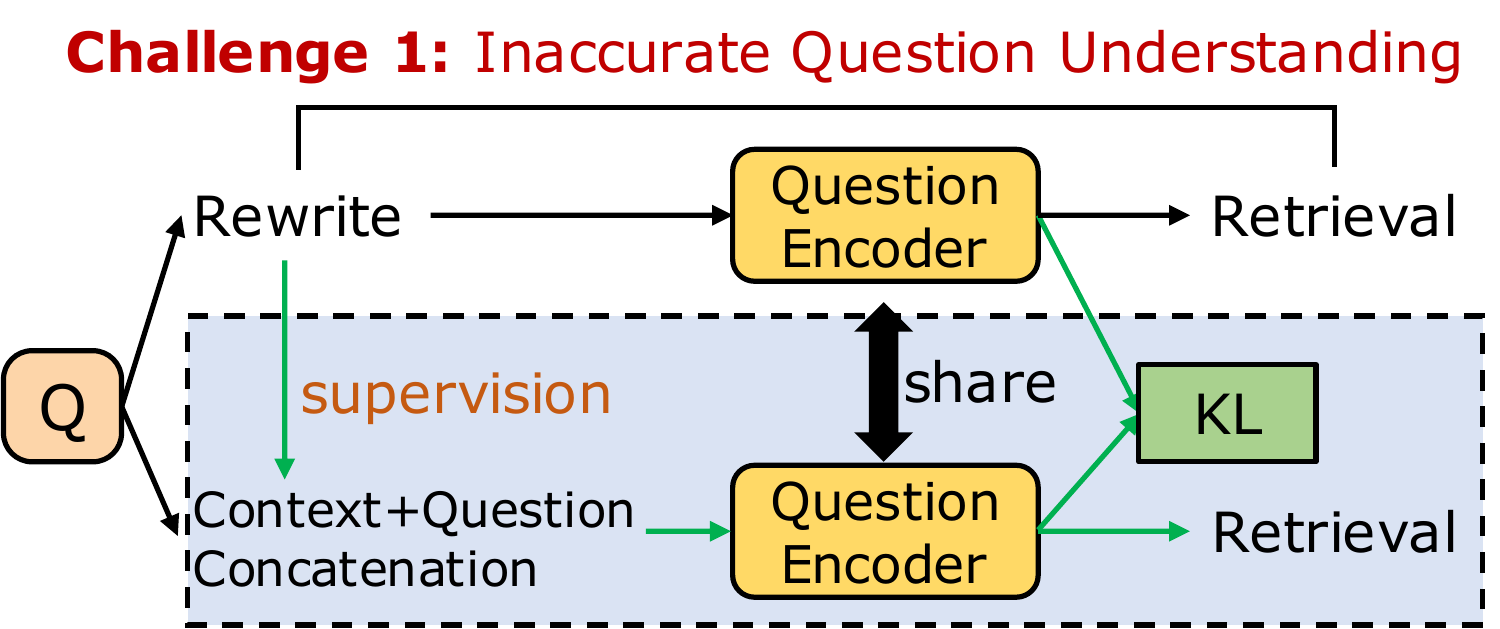}
         \caption{}
         \label{fig:limit1}
     \end{subfigure}
     \hfill
     \begin{subfigure}[b]{0.33\textwidth}
         \centering
         \includegraphics[width=\textwidth]{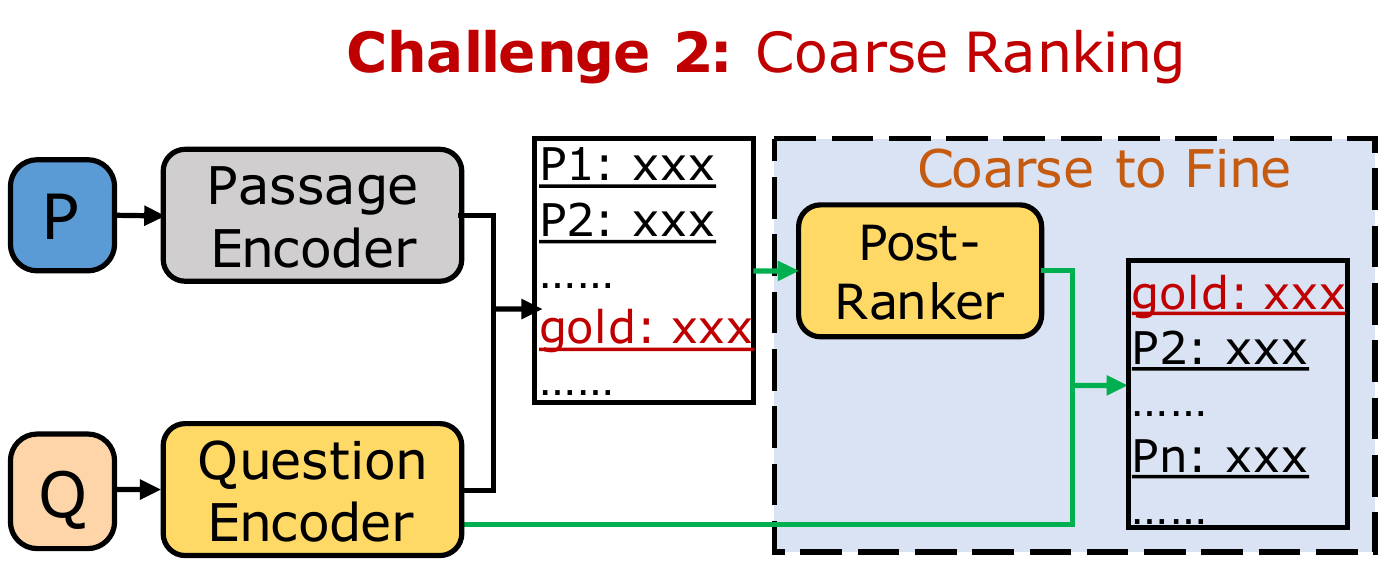}
         \caption{}
         \label{fig:limit2}
     \end{subfigure}
     \hfill
     \begin{subfigure}[b]{0.3\textwidth}
         \centering
         \includegraphics[width=\textwidth]{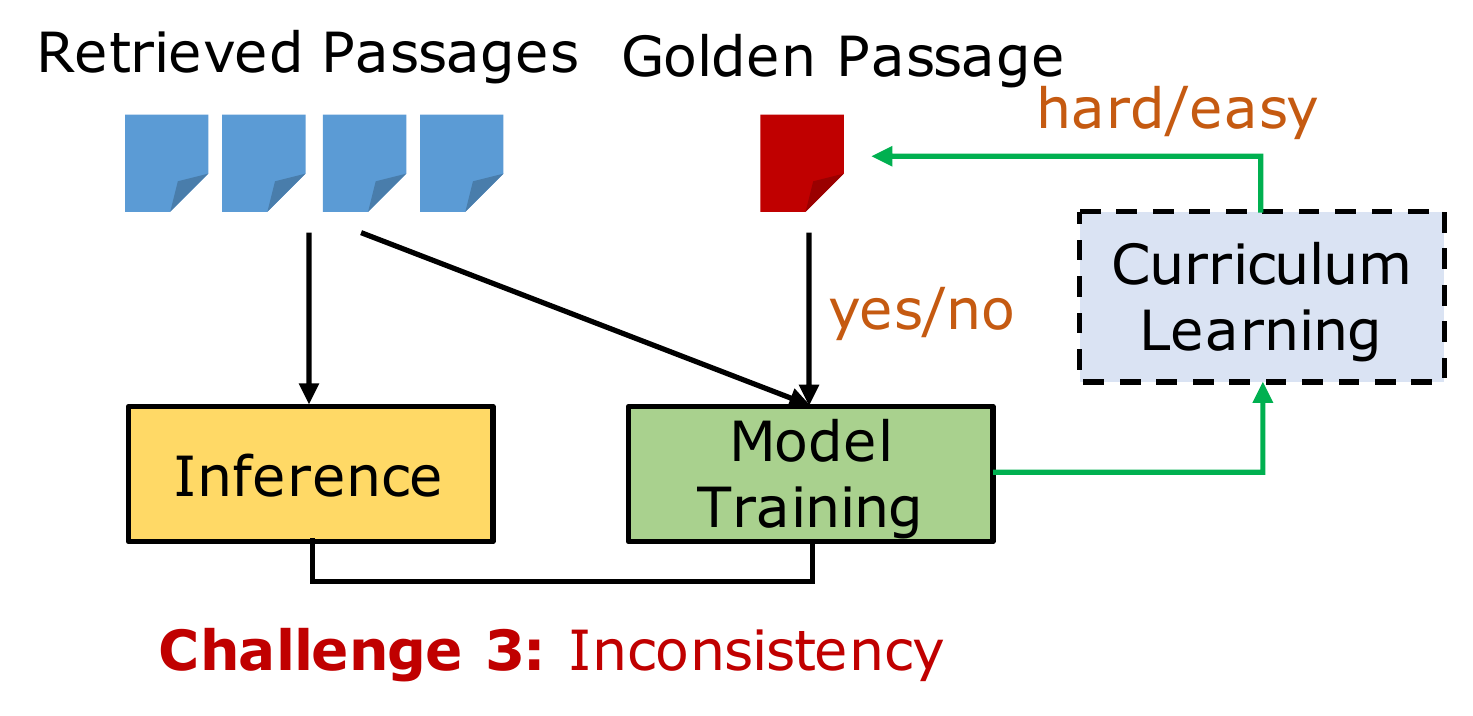}
         \caption{}
         \label{fig:limit3}
     \end{subfigure}
     \vspace{-0.15in}
        \caption{Illustration of three limitations and the corresponding solutions (marked with the blue dashed boxes). (a) Inaccurate question understanding: Feeding rewrite to the encoder might lead to poor performance in question understanding as the encoder can not learn how to model the concatenated question. (b) Coarse ranking: The relevant passage may not be picked up even it is retrieved in the initial ranking list which is roughly obtained by a frozen passage encoder and learnable question encoder. (c) Inconsistent usage of golden passage: invariably adding golden passage for training would increase dependency on it and be adverse to inference where no golden passage is manually added.}
        \label{fig:limitation}
        \vspace{-0.1in}
\end{figure*}

Open-domain question answering (OpenQA) aims to discover answers from an extremely large text source such as Wikipedia for given questions~\cite{chen2017reading, yang2019end, zhu2021retrieving}. OpenQA that receives the more widespread application as its setting is more aligned with real-world QA process of human beings. Generally, an OpenQA system needs to firstly locate a small collection of relevant articles and then generate the answer. 
The widely adopted architecture of the existing OpenQA system consists of two components, retriever and reader. The former acts as an information retrieval system to find relevant passages probably with the correct answer contained~\cite{kratzwald2018adaptive, lee2019latent,xiong2020answering,min2019knowledge,izacard2021leveraging}. The latter aims at extracting or generating the answers from the retrieved passages.
Although OpenQA enjoys a sound development momentum, it is challenged by several issues including question complexity and ambiguity, as well as insufficient background knowledge.
Conversational OpenQA which executes question answering with open-domain sources under the conversational setting is able to address the above issues by providing the context information in the conversations. 

Previous work~\cite{qu2020open} firstly defines the task of open-retrieval conversational question answering (ORConvQA) and proposes an effective end-to-end system to deal with it. 
Promising as it might be, several fundamental limitations still exist:
\textbf{(1) Inaccurate question understanding.} 
The questions in the dialogues might be hard to understand due to the unspecific pronouns. 
To alleviate this issue, ORConvQA uses rewrite questions by replacing pronouns with real entities to pre-train the retriever. Then, they concatenate the dialogue history and the question together to jointly train the whole model. However, the input of the pre-traing and the joint training process is still often inconsistent which makes the model incapable of really understand those unclear questions.
\textbf{(2) Coarse ranking.} 
Previous works adopt the \textcolor{black}{retriever-reader pipeline} which consists of a passage retriever and answer reader with a reranker/selector as regularization~\cite{yang2019end,karpukhin2020dense,qu2020open}. 
The reranker/selector here does not influence the retriever and the reader only takes the top-ranked passages as the input.
This makes the whole pipeline suffer from the coarse ranking, 
especially for the situations when golden passages are not retrieved in the top-ranked passages.
\textbf{(3) Inconsistent usage of golden passage.} 
During training, ORConvQA adds the golden passage into the training manually when the golden passage is not retrieved in the top-ranked passages. This is impractical for the testing as the golden passages are unknown in the inference stage. The inconsistency usages of golden passage during training and testing might lead to poor performance when the golden passage cannot be accurately retrieved.

To address the previous issues, we propose a framework which has Multifaceted Improvements for Conversational open-domain Question Answering ({\modelname}). {\modelname} improves the previous work from multiple aspects including regularizing retriever pre-training, as well as incorporating post-ranking and curriculum learning.
First, Figure~\ref{fig:limitation}(a) shows that, a Kullback\_Leibler (KL)-divergence based regularization is proposed for retriever pre-training, which constrains the retrieved results outputted by feeding different forms of questions. In this manner, we can learn a better representation for the concatenated question by taking its rewrite as supervision and minimizing the information loss between them. The better question understanding would benefit the retrieval and question answering performance in different phases.
Second, 
to further improve the initial retrieval results, we apply a multi-stage pipeline by adding a post-ranker module as shown in Figure~\ref{fig:limitation}(b), which also known as the post-processing~\cite{zhu2021retrieving} over retrieved passages generated by the retriever. The post-ranker learning benefits from a two-aspect constrain posed by a ranking loss with a distance-based contrastive loss. In this way, the relevant passages would be enforced by post-ranker to appear at the top positions and selected for answer reading.
Third, in the joint training stage, we design a semi-automatic curriculum learning strategy (the blue dashed boxes in Figure~\ref{fig:limitation}(c)) to reduce the dependency on the golden passages. It encourages the model to find true answer spans even when the golden passage are not retrieved as the top-ranked passages.
We evaluate {\modelname} on the publicly released dataset OR-QuAC~\cite{qu2020open} to demonstrate its effectiveness.

Our main contributions are as follows:
\begin{itemize}
    \item We propose a KL-divergence based regularization for retriever pre-training, which greatly strengthens the capability of question understanding.
    \item We propose a multi-stage pipeline for the conversational OpenQA task by incorporating a post-ranker module in the conventional retriever-reader pipline.
    \item We design a curriculum learning strategy for the joint training of the retriever, post-ranker, and reader, which effectively closes the gap between training and inference. 
    \item We conduct extensive experiments on the OR-QuAC dataset to demonstrate the effectiveness of the proposed {\modelname}. We further provide in-depth analysis on model ablation and case study to explore the insights of {\modelname}.
\end{itemize}


\begin{figure*}[t]
    \centering
    \includegraphics[width=\linewidth]{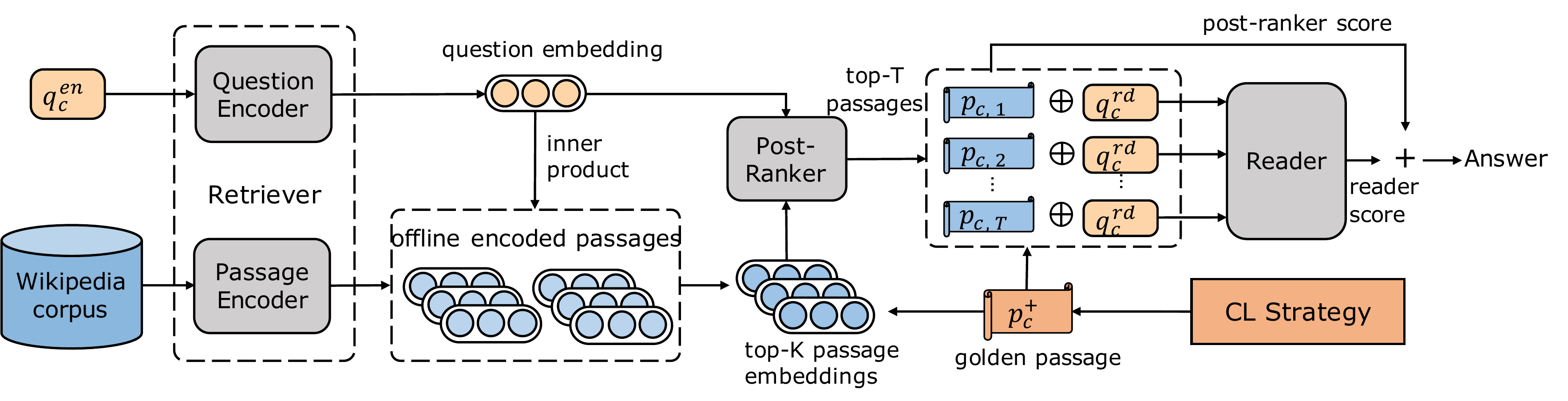}
    \caption{The overall framework of the {\modelname} model. The retriever equipped with one question encoder and one passage encoder searches the top $K$ passages from the large corpus of Wikipedia for the given question reformatted as $q_i^{en}$. The representations of $K$ passages are passed into the post-ranker to produce a refined ranking list of passages, of which top $T$ are concatenated with the question reconstructed as $q_i^{rd}$ to be the input for the reader. The reader is used to extract the final answer according to the post-ranker score and reader score.}
    \label{fig:model}
\end{figure*}

\section{Preliminary}\label{sec:preliminary}








\subsection{Retriever-Reader Pipeline}
The most typical OpenQA system follows a two-stage pipeline, which has two components: one retriever and one reader. 

\subsubsection{Retriever}\label{subsec:Retrieving and Reading}
In the retrieval process, the commonly used dual-encoder model \cite{bromley1993signature} consists of a question encoder and a passage encoder, which encode the question and passage into low-dimensional vectors, respectively. Many similarity functions, such as inner product and Euclidean distance, can be used to measure the relationship between questions and passages. The evaluation in~\cite{karpukhin2020dense} demonstrates that these functions perform comparably, so the simpler inner product is selected in our work:
\vspace{-0.05in}
\begin{equation}\label{eq:inner}
\begin{aligned}
\mathrm{sim}(q,p) = E_Q(q)^\mathrm{T}E_P(p).
\end{aligned}
\end{equation}
where $q$ and $p$ denote a given question and passage. $E_Q$ and $E_P$ refer to the question encoder and passage encoder which typically are two pre-trained model, e.g., BERT\cite{devlin2018bert}, ALBERT\cite{lan2020albert}. 
The retriever score is defined as the similarly of representations of the question and passage.
Given a question $q$, the retriever derives a small subset of passages with embeddings closest to $q$ from a large corpus.

\subsubsection{Reader}\label{subsec:Retrieving and Reading}
In reading phase, a conventional BERT-based extractive machine comprehension model is usually used as the reader. The retrieved passages concatenated with the corresponding questions are first encoded separately. 
The reader then extracts the start and end tokens with the max probabilities among tokens from all the the top passages.
The reader score is defined as the sum of the scores of start token and end token. The answer score is the sum of retriever score and reader score.
Taking the small collection of passages generated by the retriever as input, the reader outputs the answer span with the highest answer score.

\vspace{-0.05in}
\subsection{Problem Definition}
Following the task definition in ~\cite{qu2020open}, the conversational OpenQA problem can be formulated as follows: Given the current question $q_c$, and a set of historical question-answer pairs $\{(q_i, a_i)\}^{c-1}_{i=1}$, where $q_i$ denotes the $i$-th question and $a_i$ is the related answer in a conversation, the task is to identify answer spans for the current question $q_c$ from a large corpus of articles, \emph{e.g.,} Wikipedia.

Based on the introduced two-stage pipeline, this work applies a multi-stage pipeline of one pretrained retriever, one post-ranker, and one reader. The details of the framework and training process of {\modelname} are introduced in the next section.

\section{Model}\label{sec:model}
The overall framework of our {\modelname} is shown in Figure~\ref{fig:model}.
{\modelname} mainly consists of three components: (1) one regularized \textcolor{black}{retriever} for relevant passages discovery, (2) one \textcolor{black}{contrastive {\rerankname} for improving passage retrieval quality}, and (3) one \textcolor{black}{reader} for detecting answers from a small collection of ranked passages. These three components work as follows: With the passage representations offline encoded by the pre-trained passage encoder, the retriever outputs the top $K$ passages by operating inner product between them and the embedding of a given question generated by the question encoder. The post-ranker takes the initially retrieved top $K$ passage embeddings and question embedding as input, and selects the top $T$ of the newly ranked passages as output. The outputted passages are respectively concatenated with the given question, and the concatenations are fed into the reader to find the final answer. 
The answer score is decided by the post-ranker score and reader score, which would be introduced in details in Section~\ref{subsec:inference}. 
The retriever is pre-trained with a KL-divergence based regularization. Then the passage encoder of it is frozen and the question encoder can be fine-tuned with the other two components in the next joint training process in an end-to-end manner.
In the following part, we would describe the procedures of pre-training, joint training, and inference, along with the details of each component.


\subsection{KL-Divergence based Regularization for Retriever Pre-training}\label{sec:kl}
In order to improve the capability of question understanding, we propose to exploit both the original questions and question rewrites in retriever pre-training. Specifically, a regularization mechanism is proposed to force two distributions of the retrieved results generated by feeding the original questions and their rewrites into the dual-encoder to be consistent with each other by minimizing the bidirectional KL divergence between them.

Concretely, given the current question $q_c$ and its historical question-answer pairs $\{q_i, a_i\}_{i=1}^{c-1}$, we form the original question by concatenating the question-answer pairs in a history window of size $w$ with the current question. Moreover, to mitigate the issue of underspecified and ambiguous initial questions, the initial question $q_1$ is invariably considered as it makes the constructed original question closer to the question rewrite. For the question encoder, the input sequence of original question can be defined as 
$q_c^{or} = \verb|[CLS]|\;q_1 \;\verb|[SEP]|\;a_1    \;\verb|[SEP]|\;q_{c-w}\verb|[SEP]|\;a_{c-w}\;\verb|[SEP]| \cdots \verb|[SEP]|\;q_{c-1}\\
\verb|[SEP]|\;a_{c-1}\;\verb|[SEP]|\;q_c\;\verb|[SEP]|$.

\begin{figure}[t]
    \centering
    \includegraphics[width=\linewidth]{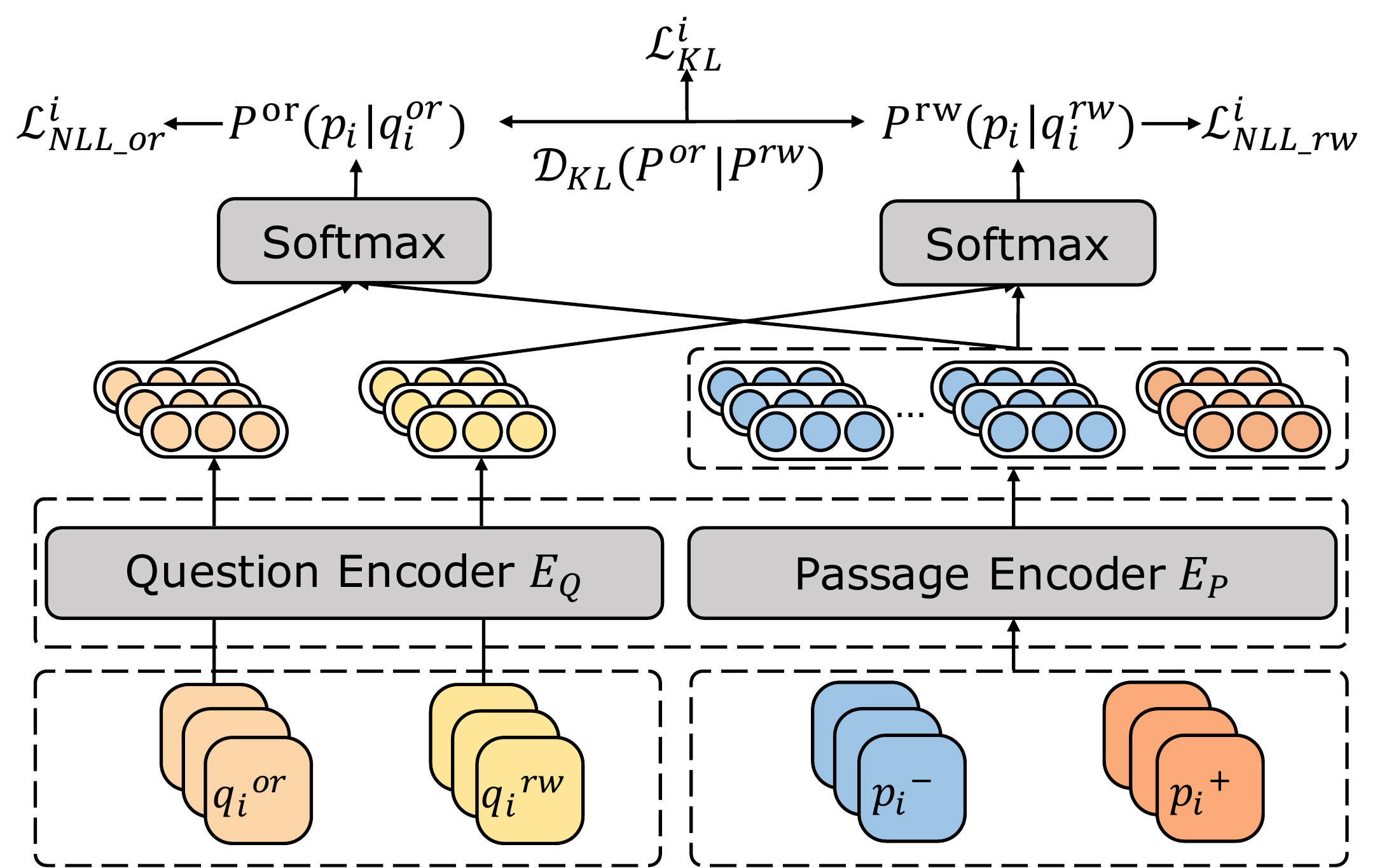}
    \vspace{-0.2in}
    \caption{Retriever pre-training model with KL-divergence based regularization. Two forms of one question are fed into question encoder while the related positive and hard negative are fed into passage encoder. The positive or negative passages of the other questions in the same batch are used as negatives.}
    \label{fig:kl}
    \vspace{-0.1in}
\end{figure}

Let $\mathcal{D} = \{(q_i^{or}, q_i^{rw}, p_i^+, p_{i,1}^-, ... ,  p_{i,n}^-)\}_{i=1}^m$ denote the training data that consists of $m$ instances. Each instance contains two forms of one question (\emph{i.e.,} original question and question rewrite), one positive passage, along with $n$ irrelevant passages $\{p_{i,j}^-\}_{j=1}^n$. These irrelevant passages used for training contain one hard negative of the given question and $n-1$ in-batch negatives which are the positive and negative samples of the other questions from the same mini-batch.
As shown in Figure~\ref{fig:kl}, we feed the original question $q_i^{or}$ and question rewrite $q_i^{rw}$ to the question encoder $E_Q$, respectively. The derived question embeddings are matched with the passage embeddings outputted by the passage encoder $E_P$ via inner product.
We can obtain two distributions of the retrieved results through softmax operation, denoted as $P^{or}(p_i|q_i^{or})$ and $P^{rw}(p_i|q_i^{rw})$. In the retrieval pre-training phase, we try to regularize on the retrieved results by minimizing the bidirectional KL divergence between the two distributions,
which can be formulated as follows:
\begin{small}
\begin{equation}
\begin{aligned}
\mathcal{L}_{KL}^i 
= & \frac{1}{2}(\mathcal{D}_{KL}(P^{or}(p_i|q_i^{or}) || P^{rw}(p_i|q_i^{rw})) \\
& + \mathcal{D}_{KL}(P^{rw}(p_i|q_i^{rw}) || P^{or}(p_i|q_i^{or}))).
\end{aligned}
\end{equation}
\end{small}

For the main task of passage retrieval, we apply the widely used negative log likelilhood of the positive passage given two forms of one question as objective function:
\begin{small}
\begin{equation}
\begin{aligned}
\mathcal{L}_{NLL}^i 
= & - \frac{1}{2}(\mathcal{L}_{NLL\_or}^i+\mathcal{L}_{NLL\_rw}^i)\\
=&- \frac{1}{2}(\mathrm{log}P^{or}(p_i^+|q_i^{or}) + \mathrm{log}P^{rw}(p_i^+|q_i^{rw})),
\end{aligned}
\end{equation}
\end{small}
\\where the probability of retrieving the positive passage can be calculated as:
\begin{small}
\begin{equation}
 P^{or}(p_i^+|q_i^{or}) 
 = \frac{\mathrm{exp}(\mathrm{sim}(q_i^{or}, p_i^+))}{\mathrm{exp}(\mathrm{sim}(q_i^{or}, p_i^+))+\sum_{j=1}^n \mathrm{exp}(\mathrm{sim}(q_i^{or}, p_{i,j}^-))},
\end{equation}
\end{small}
\begin{small}
\begin{equation}
 P^{rw}(p_i^+|q_i^{rw})
 = \frac{\mathrm{exp}(\mathrm{sim}(q_i^{rw}, p_i^+))}{\mathrm{exp}(\mathrm{sim}(q_i^{rw}, p_i^+))+\sum_{j=1}^n \mathrm{exp}(\mathrm{sim}(q_i^{rw}, p_{i,j}^-))}.
\end{equation}
\end{small}
\\The probability of retrieving those negative passages can be obtained in the same manner.

To pre-train the retriever, 
the training objective is to minimize the pre-retrieval loss $\mathcal{L}^i_{pre}$ for data ($q_i^{or}$, $q_i^{rw}$, $p_i^+$, $p_{i,1}^-$, ... ,  $p_{i,n}^-)$:
\begin{equation}
\mathcal{L}^i_{pre} = \mathcal{L}_{NLL}^i + \alpha \mathcal{L}_{KL}^i,
\end{equation}
where $\alpha \in [0, 1]$ is a hyperparameter used  to control $\mathcal{L}_{KL}^i$.

\subsection{Joint Training with Curriculum Learning}
With the passage encoder and question encoder of the retriever pre-trained, our {\modelname} jointly trains the question encoder, post-ranker, and reader with a designed curriculum learning strategy.

\subsubsection{Retriever Loss}\label{subsec:retriever}
The pre-trained passage encoder is used to embed the open-domain passages offline and obtain a set of passage representations. 
Given the current question $q_c$, and its historical question-answer pairs $\{q_i, a_i\}_{i=c-w}^{c-1}$ within a window size $w$, the input for question encoder is constructed by concatenating the question-answer pairs with the current question as mentioned in Section~\ref{sec:kl}. With the prepared offline passage representations and outputs of question encoder, each passages would be assigned with a retrieval score computed by the inner product operation as shown in Equation~\ref{eq:inner}. We select the top $K$ passages with high retrieval scores for the \textcolor{black}{{\rerankname} and reader}.

For each question, the reformatted question for the question encoder is denoted as $q_i^{en}$.
The question encoder of the retriever is fine-tuned by optimizing the following retrieval loss:
\begin{equation}\label{eq:retriever}
    \mathcal{L}^i_{retriever}=-\mathrm{log}\frac{\mathrm{exp}(\mathrm{sim}(q_i^{en},p_i^+))}{\sum_{j=1}^{K}\mathrm{exp}(\mathrm{sim}(q_i^{en},p_{i,j}))},
\end{equation}
where $p_{i,j}$ denotes the retrieved passages regarding question $q_i$.

For a fair comparison, we omit the answers from the concatenated question in the experiment to keep the same setting with the previous work~\cite{qu2020open}.
\subsubsection{Post-Ranker with Contrastive Loss}
Since only the first $T$ of the top $K$ passages outputted by the retriever module are fed into the reader for answer extraction, a {\rerankname} is proposed to re-rank the $K$ passages to push more relevant passages be ranked in the top $T$ placement.
We build the {\rerankname} by adding a subsequent network after the pre-trained passage encoder, which takes the embeddings of $K$ passages as input. With the subsequent network, {\rerankname} is able to learn high-level feature representations for passages.
In this work, we use the linear layer as the subsequent network for simplicity. The output for the retrieved passage $p_{i,j}$ and the ranking score can be formulated as follows:
\vspace{-0.1in}
\begin{equation}
    \mathbf{d}_{i,j} = \mathrm{LinearLayer}(E_P(p_{i,j})),
\end{equation}
\begin{equation}\label{eq:score_post}
    S_{post}(q_i^{en},p_{i,j})=\mathbf{d}_{q_i}^{\mathrm{T}}\mathbf{d}_{i,j},
\end{equation}
where $\mathbf{d}_{q_i}$ is the representation for question $q_i$ generated by the question encoder $E_Q$. $S_{post}(\cdot)$ refers to the scores generated by the {\rerankname} network.

To simultaneously fine-tune the question encoder and {\rerankname}, we apply the modified hinge loss combined with a distance-based contrastive loss to pose constraints for passage reranking from two aspects. The hinge loss function~\cite{santos2016attentive} is defined as: 
\begin{small}
\begin{equation}\label{eq:rerank}
    \mathcal{L}^i_{ranker}=\mathrm{max}\{0,\delta-S_{post}(q_i^{en},p_i^+)+\mathrm{max}_j\{S_{post}(q_i^{en},p_{i,j}^-)\}\},
\end{equation}
\end{small}
where $\delta$ is the margin of the hinge loss, $p_i^+$ and $p_{i,j}^-$ denote the positive passage and negative passages retrieved for $q_i$. 
For contrastive learning, we uses the triplet margin loss~\cite{weinberger2006distance} to measure the distance between positive and negative samples as follows: 
\vspace{-0.05in}
\begin{equation}\label{eq:contrastive}
    \mathcal{L}^i_{cl}=\mathrm{max}\{0,\mu+D(\mathbf{d}_{q_i},\mathbf{d}^+_i)-D(\mathbf{d}_{q_i},\mathbf{d}_{i,j}^-)\},
    \vspace{-0.05in}
\end{equation}
where $\mu$ is the margin of the triplet margin loss. We apply Euclidean distance $D$ to compute the distance between the question and passage in the representation space. The final {\rerankname} loss is defined as:
\vspace{-0.1in}
\begin{equation}\label{eq:contrastive}
    \mathcal{L}^i_{postranker}= \mathcal{L}^i_{ranker}+\beta\mathcal{L}^i_{cl},
    \vspace{-0.05in}
\end{equation}
where $\beta$ is hyperparameter that need to be determined.

\subsubsection{Reader Loss}
The neural reader of our {\modelname} is designed for predicting an correct answer to the given question. Given the top $T$ retrieved/reranked passages, the reader intends to detect an answer span from each passage with a span score. Moreover, a passage selection score is assigned to each passage. The passage selection is used to select the passages that contains the true answer of the given question~\cite{lin2018denoising,karpukhin2020dense}, which performs as a contraint to facilitate the training of reader. The span with the highest passage selection score and span score is extracted as the final answer. 

Our reader applies the widely used pre-trained model, such as BERT~\cite{devlin2018bert}, RoBERTa~\cite{liu2019roberta}. Given a question $q_i$ and the top $T$ previously retrieved passages, we form the question input in the same way as that for retriever described in Section~\ref{subsec:retriever}. The only difference is that the initial question $q_1$ is left out as the setting of~\cite{qu2020open}. We denote the reconstructed question as $q_i^{rd}$ and concatenate it with a retrieved passage to form the sequence as the input for the pre-trained model. Specifically, let $\mathbf{z}_{i,j,t}$ be the outputted token-level representations of the $t$-th token in the passage $p_{i,j}$ which is retrieved for question $q_i^{rd}$. $\mathbf{z}_{i,j,[\mathrm{cls}]}$ denotes the sequence-level representation for the input sequence. The scores for the $t$-th token being the start and end tokens, and a passage being selected are defined as follows:
\begin{equation}\label{eq:score_reader1}
    S_{s}(q_{i}^{rd},p_{i,j},[t]) = \mathbf{z}_{i,j,t}^\mathrm{T}\mathbf{w}_{start},
\end{equation}
\begin{equation}\label{eq:score_reader2}
    S_{e}(q_{i}^{rd},p_{i,j},[t]) = \mathbf{z}_{i,j,t}^\mathrm{T}\mathbf{w}_{end},
\end{equation}
\begin{equation}\label{eq:score_select}
    S_{select}(q_{i}^{rd},p_{i,j})=\mathbf{z}_{i,j,[\mathrm{cls}]}^\mathrm{T}\mathbf{w}_{select},
\end{equation}
where $\mathbf{w}_{start}$, $\mathbf{w}_{end}$, and $\mathbf{w}_{select}$ are trainable vectors.

The loss function for the start token prediction is defined as:
\begin{equation}
    \mathcal{L}^i_{start}=-\mathrm{log}\frac{\mathrm{exp}(S_s(q_i^{rd},p_i^+,[st_i^+]))}{\sum_{j=1}^{M}\sum_{[t]\in p_{i,j}}\mathrm{exp}(S_s(q_i^{rd},p_{i,j}, [t]))},
\end{equation}
where $[st_i^+]$ denotes the start token of the true answer in the golden passage $p_i^+$.
The objective function for the end token prediction denoted as $\mathcal{L}^i_{end}$ is defined in the same manner. In addition, the passage selecting loss is computed as follows:
\begin{equation}
    \mathcal{L}^i_{select}=-\mathrm{log}\frac{\mathrm{exp}(S_{select}(q_i^{rd},p_i^+))}{\sum_{j=1}^{M} \mathrm{exp}(S_{select}(q_i^{rd},p_{i,j}))}.
\end{equation}
Finally, the loss function of the reader is defined as follows:
\begin{equation}\label{eq:reader}
    \mathcal{L}_{reader}^i = \frac{1}{2}(\mathcal{L}_{start}^i+\mathcal{L}_{end}^i)+\mathcal{L}_{select}^i.
\end{equation}

\subsubsection{Joint Training with Curriculum Learning Strategy}
We propose a semi-automatic curriculum learning (CL) strategy consisting of two core components: an automatic difficulty measurer and a discrete training scheduler, to improve the joint training of retriever, post-ranker, and reader.
The CL strategy aims at reducing the chance of adding golden passage during the joint training process to help the settings of training and inference be more consistent. 
It would effectively push the retriever to find the passage with the true answer contained by itself and encourage the reader to discover the correct answer without the assistance of golden passage.
To the best of our knowledge, this is the first work that designs CL strategy for conversational OpenQA joint learning.

Specifically, our semi-automatic CL strategy automates the difficulty measurer by taking the question-wise training loss as criteria. A higher retriever loss is recognized a harder mode for the retriever to discover the positive passages. For the hard mode, the training scheduler of our CL strategy introduces the golden passage in the retrieval results to make the learning easier. 
Our training scheduler is like a Baby Step scheduler~\cite{bengio2009curriculum,spitkovsky2010baby} with a finer granularity as shown in Algorithm~\ref{alg:TS}.



Formally, for each iteration $l$, let $\mathcal{P}^{(l)}_K=\{\mathcal{P}^{(l)}_i|q_i\in Q^{(l)}\}$ be the set of all the collections of top $K$ passages retrieved by the retriever to questions of the batch in the iteration. For the hard mode, the training scheduler brings the golden passage of each question to $\mathcal{P}^{(l)}_K$ to form the set of retrieved passages with the positive one, denoted as $\mathcal{P}^{(l)}_{KG}=\{\mathcal{P}^{(l)}_i\cup\{p^+_i\}|q_i\in Q^{(l)}\}$. With these retrieved passages, the final loss to be optimized is defined as:
\begin{equation}\label{eq:final}
    \mathcal{L}^{(l)}_{final} = v^{(l)}\mathcal{L}^{(l)}(\mathcal{P}_{KG}^{(l)})+(1-v^{(l)})\mathcal{L}^{(l)}(\mathcal{P}^{(l)}_{K}),
\end{equation}
where $v^{(l)}$ is determined by the difficulty measurer with the retriever loss of the previous iteration:
\begin{equation}\label{eq:v}
    v^{(l)}=\left\{
    \begin{array}{ccl}
     1,  & & {\mathcal{L}^{(l-1)}_{retriever}>\lambda_{upper}}, \\
     0,  & & {\mathcal{L}^{(l-1)}_{retriever}<\lambda_{lower}}, \\
     I(p_b), & & {\text{otherwise}},
    \end{array} \right.
\end{equation}
where $\lambda_{upper}$ and $\lambda_{lower}$ are the pre-defined upper and lower thresholds. $I(p_b)$ denotes an indicator function whose output is sampled from a Bernoulli distribution:
\begin{equation}
    I(p_b)=\left\{
    \begin{array}{ccl}
     1,  & & {\text{with probability}\ p_b}, \\
     0,  & & {\text{with probability}\ 1-p_b}, 
    \end{array} \right.
\end{equation}
where the probability is evaluated by the min-max normalization denoted as $p_b=\frac{ \mathcal{L}^{(l-1)}_{retriever}-\lambda_{lower}}{\lambda_{upper} - \lambda_{lower}}$, which measures the degree of approximating upper threshold for the retriever loss.
It can be intuitively explained that if the retrieval loss of the previous iteration is close to an upper threshold $\lambda_{upper}$, the training status is regarded in the hard mode.
And the value of $v^{(l)}$ is assigned with 1 in a higher probability to select the set of retrieved passages with golden passage for training in the current iteration.

$\mathcal{L}^{(l)}(\cdot)$ in Equation~\ref{eq:final} denotes the objective function integrated by all the three modules with the retrieved passages in the $l$-th iteration as follows:
\begin{equation}
    \mathcal{L}^{(l)} = \mathcal{L}^{(l)}_{retriever} + \mathcal{L}^{(l)}_{postranker}+ \mathcal{L}^{(l)}_{reader}.
\end{equation}
The loss of each module is averaged over the questions in the $l$-th iteration.
\SetKwComment{Comment}{/* }{ */}

\begin{algorithm}[t]
\caption{Training Scheduler}\label{alg:TS}
\LinesNumbered
\KwIn{Training data $\mathcal{D}_{train}=\{(q_i, a_i, p_i^+)\}_{i=1}^m$, \\
\qquad \quad Iteration number $L$.}
\KwOut{A set of optimal model parameters.}

\For{$l=1,\cdots, L$}{
    Sample a batch of questions $Q^{(l)}$\\
    \For{$q_i\in Q^{(l)}$}{
        $\mathcal{P}_{i}^{(l)} \gets \mathrm{arg\,max}_{p_{i,j}}(\mathrm{sim}(q_i^{en},p_{i,j}),K)$\\
        $\mathcal{P}_{Gi}^{(l)} \gets \mathcal{P}_{i}^{(l)}\cup\{p^+_i\}$\\
        Compute $\mathcal{L}^i_{retriever}$, $\mathcal{L}^i_{postranker}$, $\mathcal{L}^i_{reader}$\\ according to Eq.\ref{eq:retriever}, Eq.\ref{eq:rerank}, Eq.\ref{eq:reader}\\
    }
    $\mathcal{L}^{(l)} \gets \frac{1}{|Q^{(l)}|}\sum_i(\mathcal{L}^{i}_{retriever} + \mathcal{L}^{i}_{postranker}+ \mathcal{L}^{i}_{reader})$\\
    $\mathcal{P}^{(l)}_K\gets\{\mathcal{P}^{(l)}_i|q_i\in Q^{(l)}\}$,\quad $\mathcal{P}^{(l)}_{KG}\gets\{\mathcal{P}^{(l)}_{Gi}|q_i\in Q^{(l)}\}$\\
    Compute the coefficient $v^{(l)}$ according to Eq.~\ref{eq:v}\\
  \eIf{$ v^{(l)}=1$}{
    $\mathcal{L}^{(l)}_{final} \gets \mathcal{L}^{(l)}(\mathcal{P}_{KG}^{(l)})$\\
  }{
      $\mathcal{L}^{(l)}_{final} \gets \mathcal{L}^{(l)}(\mathcal{P}^{(l)}_{K}),$\\
    }
    Optimize $\mathcal{L}^{(l)}_{final}$
}
\end{algorithm}



   
\subsection{Inference}\label{subsec:inference}
In the inference stage, for a given question $q_c$ and its historical question-answer pairs $\{q_i, a_i\}_{i=c-w}^{c-1}$ within a window size $w$, we first obtain a collection of the top $T$ relevant passages through the two consecutive modules, namely retriever and {\rerankname}. For each passage $p_{c,j}$ in the collection, we get the {\rerankname} score $S_{post}(q_c^{en},p_{c,j})$ according to Equation~\ref{eq:score_post}. The reader score includes two parts, one is the select score $S_{select}(q_{c}^{rd},p_{c,j})$ calculated by Equation~\ref{eq:score_select}, the other one is the span score which consists of the start score and end score as shown in Equations~\ref{eq:score_reader1} and~\ref{eq:score_reader2}. The span score can be obtained by:
\begin{footnotesize}
\begin{equation}
    S_{span}(q_{c}^{rd},p_{c,j},sp)=\max\limits_{sp\in p_{c,j}}\{ S_{s}(q_{c}^{rd},p_{c,j},[t_s])+ S_{e}(q_{c}^{rd},p_{c,j},[t_e])\},
\end{equation}
\end{footnotesize}
where $sp=([t_s],[t_e])$ is the answer span starting with token $[t_s]$ and ending with token $[t_e]$. Following the previous work~\cite{kenton2019bert,qu2020open},  we pick out the top 20 spans and discard the invalid predictions including the cases where the start token comes after the end token, or the predicted span overlaps with the question part of the input sequence. The final prediction score is defined as follows:
\begin{equation}
    S(q_{c}^{en},q_{c}^{rd},p_{c,j},sp) = S_{post}(q_c^{en},p_{c,j}) + S_{reader}(q_{c}^{rd},p_{c,j}, sp),
\end{equation}
where $S_{reader}(q_{c}^{rd},p_{c,j},sp) = S_{select}(q_{c}^{rd},p_{c,j}) + S_{span}(q_{c}^{rd},p_{c,j},sp)$. For the given question $q_c$ in a conversation, the answer span $sp$ in the retrieved passage $p_{c,j}$ that has the largest score is the predicted answer. 
\section{Experiments}\label{sec:exp}
\subsection{Dataset}
To evaluate the effectiveness of the proposed {\modelname}, we conduct comprehensive experiments on the public available dataset OR-QuAC~\cite{qu2020open}, which integrates three datasets including the QuAC~\cite{choi2018quac}, CANARD~\cite{elgohary2019can}, and the Wikipedia corpus.
OR-QuAC consists of totally 5,644 conversations containing 40,527 questions and answers with the rewrites of questions obtained from CANARD. The question rewrites support the \textcolor{black}{KL divergence-based regularization} for pretraining a better passage retriever. 
OR-QuAC also provides a collection of more than 11 million passages obtained from the English Wikipedia dump from 10/20/2019\footnote{ttps://dumps.wikimedia.org/enwiki/20191020/} for open-retrieval.
Table \ref{tab:dataset} summarizes the statistics of the aggregated OR-QuAC dataset. 


\begin{table}[t]\label{tab:data}
  \centering
  \caption{Data statistics of the OR-QuAC dataset.}
    \begin{tabular}{llccc}
    \toprule
          & Items & Train & Dev   & Test \\
    \midrule
    \multirow{5}[2]{*}{Coversations} & \# Dialogs & 4,383 & 490   & 771 \\
          & \# Questions / Rewrites & 31,526 & 3430  & 5571 \\
          & \# Avg. Questions / Dialog & 7.2   & 7     & 7.2 \\
          & \# Avg. Tokens / Question & 6.7   & 6.6   & 6.7 \\
          & \# Avg. Tokens / Rewrite & 10    & 10    & 9.8 \\
    \midrule
    Wikipedia  & \# Passages & \multicolumn{3}{c}{11 million} \\
    \bottomrule
    \end{tabular}%
  \label{tab:dataset}%
\end{table}%

\renewcommand{\arraystretch}{1.2}
\begin{table*}[t]
  \centering
  \caption{Performance comparison of {\modelname} and baseline models. The number in the parentheses is the batch size during the retriever pre-training.}
  \vspace{-0.1in}
     \begin{tabular}{p{10.5em}cccccccccc}
    \toprule
    \multirow{2}[4]{*}{Methods} & \multicolumn{5}{c}{Development}                       & \multicolumn{5}{c}{Test} \\
\cmidrule(lr){2-6} \cmidrule(lr){7-11}     \multicolumn{1}{c}{} & F1    & HEQ-Q & HEQ-D & Rt MRR & Rt Recall & F1    & HEQ-Q & HEQ-D & Rt MRR  & Rt Recall \\
    \midrule
    DrQA~\cite{chen2017reading}  & 4.5   & 0.0   & 0.0   & 0.1151   & 0.2000  & 6.3   & 0.1   & 0.0   & 0.1574   & 0.2253 \\
    BERTserini~\cite{yang2019end} & 19.3  & 14.1  & 0.2   & 0.1767  & 0.2656  & 26.0  & 20.4  & 0.1   & 0.1784   & 0.2507 \\
    DPR (16)~\cite{karpukhin2020dense} & 25.9  & 16.4  & 0.2   & 0.3993  & 0.5440  & 26.4  & 21.3  & 0.5   & 0.1739  & 0.2447  \\
    \midrule
    ORConvQA-bert (64)~\cite{qu2020open} & 26.9  & 17.5  & 0.2   & 0.4286   & 0.5714  & 29.4  & 24.1  & 0.6   & 0.2246  & 0.3141  \\
    ORConvQA-roberta (64) & 26.5 & 17.8 & 0.2 &	0.4284  & 0.5624 & 	28.7 &	24.2 &	0.8 & 0.2330  & 0.3226 \\
    \midrule
    ours-bert (16)  & 28.0 &	19.4 &	0.2 & 0.4639  & 0.6157 & 31.7 & 27.8 &	1.2 & 0.2763  & 0.3668  \\
    ours-roberta (16) & 28.1 & 19.5 & \textbf{0.4} & 0.4639 & 0.6169 & 33.4 & 29.4 & 1.7 & 0.2887  & 0.3819\\
    ours-bert (32) & 27.6 &	19.6 &	0.0 & \textbf{0.4675}  & 0.6236 & 32.6 & 29.1 & 0.8 & 0.3013  & 0.4130  \\
    ours-roberta (32)& \textbf{29.4}  & \textbf{20.2}  & \textbf{0.4}   & 0.4656   & \textbf{0.6248}  & \textbf{35.0}  & \textbf{30.8}  & \textbf{1.8}   & \textbf{0.3073}  & \textbf{0.4202}  \\
    \bottomrule
    \end{tabular}%
  \label{tab:overall}%
\end{table*}%

\vspace{-0.1in}
\subsection{Experimental Settings}

\subsubsection{Evaluation Metrics}
Following the evaluation protocols used in~\cite{qu2020open}, we apply the word-level F1 and the human equivalence score (HEQ) that are provided by the QuAC challenge~\cite{choi2018quac} to evaluate our {\modelname}. As the core evaluation metric for the overall performance of answer retrieval, F1 is computed by considering the portion of words in the prediction and groud truth that overlap after removing stopwords.
HEQ is used to judge whether a system's output is as good as that of an average human. It measures the percentage of examples for which system F1 exceeds or matches human F1. Here two variants are considered: the percentage of questions for which this is true (HEQ-Q), and the percentage of dialogs for which this is true for every questions in the dialog (HEQ-D).
Furthermore, we use another two metrics, Mean Reciprocal Rank (MRR) and Recall for the evaluation of the retrieval performance. MRR reflects the abilities of post-ranker to return the passages containing true answers in a high place.
Recall is indicative of post-ranker's capability of providing relevant passages for the next modules. 
For the sake of fairness, we follow ~\cite{qu2020open} and calculate the two metrics for the top $T$ passages that are retrieved for the reader.


\vspace{-0.1in}
\subsubsection{Implementation Settings}
\begin{enumerate}
    \item \textbf{Retriever Pretraining}. As question and passage encoders, we employ two ALBERT Base models. The maximum sequence length for the question encoder is 128, while the maximum length for the passage encoder is 384. The models are trained on NVIDIA TITAN X GPU. The training batch size is set to 16, the number of training epochs is set to 12, the learning rate is set to 5e-5, the window size $w$ is set to 6 and the coefficient of KL divergence $\alpha$ is set to 0.2. Every 5,000 steps, we save checkpoints and assess on the development set provided by \cite{qu2020open}. Then we select several models that performed well on the development set and apply them on test questions and wiki passages. Finally, we select the best model for future training.
    \item \textbf{Post-ranker and Reader}. For the post-ranker, we utilize a linear layer with a size of 128 for simplicity. The number $K$ of passage embeddings it takes as input is set to 100, and the number $T$ of its outputted passages for reader is set to 5. For the reader, we apply the BERT and RoBERTa model. The max sequence length is assigned with 512. The sequence is concatenated by a question and a passage. The maximum passage length is set to 384, with the remainder reserved for the question and other tokens such as  \verb|[CLS]| and  \verb|[SEP]|.
    \item \textbf{Joint Training}. We use the pre-trained passage encoder to compute an embedding for each passage, and build a single MIPS index using FAISS\cite{JDH17} for fast retrieval. Models are jointly trained on NVIDIA TITAN X GPU. 
    The training batch is set to 2, the number of epochs is 3, the learning rate is 5e-5, and the optimizer is Adam.
    The hyper-parameters including the margin of hinge loss $\delta$, triplet margin loss $\mu$, coefficient of contrastive loss $\beta$, and pre-defined threshold $\lambda_{upper}$ and $\lambda_{lower}$ are tuned based on the development set to select the optimal model for inference.
    We save checkpoints and evaluate on the development set every 5,000 steps, and then select the best model for the test set.
\end{enumerate}

\vspace{-0.1in}
\subsubsection{Baselines}
To demonstrate the effectiveness of our proposed {\modelname}, we compare it with the following state-of-the-art question answering models, including three representative OpenQA models (DrQA, BERTserini, and DPR) and one conversational OpenQA model (ORConvQA):
\begin{itemize}
    \item \textbf{DrQA}~\cite{chen2017reading} is composed of a document retriever which uses bigram hashing and TF-IDF matching to return the relevant passages for a given question, and a multi-layer RNN based document reader for answer spans detection in those retrieved passages.
    \item \textbf{BERTserini}~\cite{yang2019end} uses a BM25 retriever from Anserini\footnote{http://anserini.io/} and a BERT reader to tackle end-to-end question answering. The retriever directly identifies segments of open-domain texts and pass them to the reader. Compared to DPR, ORConvQA and our {\modelname}, it has no selection loss in the reader and benefits less from the joint learning.
    \item \textbf{DPR}~\cite{karpukhin2020dense} increases retrieval by learning dense representations instead of using typical IR methods. It innovatively proposes to introduce hard negatives in the training process of the retriever. For OpenQA, DPR consists of a dual encoder as a retriever and BERT as a reader. 
    \item \textbf{ORConvQA}~\cite{qu2020open} is first proposed to solve the conversational open-domain QA problem with the retriever, rerank, and reader pipeline. This is the key work to compare for our {\modelname}. 
\end{itemize}

The results of all the baselines except DPR come from the previous work~\cite{qu2020open}. The implementation setting of DPR is the same with ORConvQA, including the encoder network, window size, learning rate, and so on.

\vspace{-0.1in}
\subsection{Overall Results}
The overall experimental results are reported in Table~\ref{tab:overall}. The results of the baseline models are public in~\cite{qu2020open} except for DPR. The retrieval metrics denoted by ``Rt MRR'', and ``Rt Recall'' are used to evaluate the retrieval results of retriever for baselines while evaluating that of post-ranker for {\modelname}.
Generally, our {\modelname} outperforms all the baseline models. In detail, several observations can be achieved:
\begin{enumerate}
    \item Both DrQA and BERTserini achieve poor performance. The primary reason is they use the sparse retriever which can not be fine-tuned in the downstream  reader training to discover relevant passages for answer extraction. DrQA performs rather badly in answer reading as it uses RNN-based reader which does not have the strong ability of representation learning as those pre-trained language model. 
    The reader of BERTserini is similar to the other compared methods except DrQA. But it benefits less from the multi-task training process as there is no select component (reranker) in reader. 
    Compared with DrQA and BERTserini, DPR improves the retriever with the dense representation learning. The performance of DPR is limited by the batch size of pre-training.
    
    \item As the first system designed for the task of conversational OpenQA, ORConvQA provides the best performance among the baselines. ORConvQA is similar with DPR, where the difference is that it does not use hard negatives for retriever pre-training while DPR uses. The main reason ORConvQA performs better is the batch size of retriever pre-training is assigned with 64 as it uses 4 GPUs and set batch size to 16 per GPU. Actually, with the same batch size, the retriever of DPR is stronger, detailed results of which can be seen in the further analysis of Section~\ref{subsec:fa1}. 
    \item Our {\modelname} gives the significantly better performance than ORConvQA, even though our retriever is pre-trained with batch size of 16. This indicates that our {\modelname} can perform better with the lower memory cost. When the batch size is increased to 32, the performance can be further improved. The results convincingly demonstrate the effectiveness of the multifaceted improvements of KL-divergence regularization based pre-training, the added post-ranker, and the semi-automatic curriculum learning strategy.
    \item To investigate the influence of the pre-trained language model of reader, we evaluate our {\modelname} and ORConvQA based on two language models, namely BERT and RoBERTa. For ORConvQA, the results obtained by using BERT and RoBERTa are comparable. For our {\modelname}, applying RoBERTa for reader effectively improves the performance compared to using BERT. Overall, {\modelname} achieves the best performance whether with BERT or RoBERTa.
\end{enumerate}



\vspace{-0.15in}
\subsection{Ablation Studies}
\begin{table}[t]
  \centering
  \caption{Performance of ablation on different components. {\modelname} refers to the full system.}
  \vspace{-0.1in}
    \begin{tabular}{clllll}
    \toprule
    \multicolumn{2}{l}{Settings} & {\modelname}  & \makecell[l]{w/o KL \\retriever} & \makecell[l]{w/o \\post-ranker} & \makecell[l]{w/o \\curriculum} \\
    \midrule
    \multirow{5}[2]{*}{Dev} & F1    & \textbf{28.1} & 27.7  & 27.8  & 27.8  \\
          & HEQ-Q & \textbf{19.5} & 19.4  & 19.4  & 18.3  \\
          & HEQ-D & \textbf{0.4} & 0.0   & 0.0   & 0.0  \\
          & Rt MRR & \textbf{0.4639} & 0.4033  & 0.4604  & 0.4560  \\
          & Rt Recall & \textbf{0.6169} & 0.5324  & 0.6157  & 0.6140  \\
    \midrule
    \multirow{5}[2]{*}{Test} & F1    & \textbf{33.4} & 31.5  & 32.2  & 31.7  \\
          & HEQ-Q & \textbf{29.4} & 29.0  & 28.2  & 27.1  \\
          & HEQ-D & \textbf{1.7} & 1.4   & 1.6   & 0.8  \\
          & Rt MRR & \textbf{0.2887} & 0.2110  & 0.2810  & 0.2812  \\
          & Rt Recall & 0.3819  & 0.2968  & \textbf{0.3830} & 0.3764  \\
    \bottomrule
    \end{tabular}%
    \vspace{-0.15in}
  \label{tab:ablation}%
\end{table}%
To investigate the effectiveness of three improved parts in our {\modelname}, we evaluate several variants of our system. As shown in Table~\ref{tab:ablation}, once we remove one of the three components, both the retrieval and QA performance generally decrease. The detailed observations are summarized as follows:
\begin{enumerate}
    \item When we remove the KL-based regularized pre-trained retriever and use the retriever of ORConvQA as replacement, the performance drops significantly, especially the retrieval performance. It shows the importance of the KL-based regularization in our pre-trained retriever.
    \item Removing the post-ranker also brings a degradation in the overall performance. The influence is slightly smaller than that of KL-based regularization, which is probably caused by the simple linear layer used for post-ranker. This is our limitation and we plan to explore the more flexible and effective neural network for post-ranker in future work. From another perspective, some improvements can be achieved just by adding a linear layer for further passage representation learning.
    It is notable that the retrieval recall is higher than the full system. It is mainly because each question in the test set has more than one golden passages. Retrieving more golden passages does not necessarily lead to the better answer span which has the higher coverage of the ground truth answer.
    \item The variant without curriculum learning in the joint training performs worse than the full {\modelname}. By comparison, the QA performance decreases more noticeably. The reason behind is that the curriculum learning strategy encourages the reader to find correct answer with no golden passage assistance at the joint training time. It makes the reader more suitable for answer extraction in the inference phase.
\end{enumerate}

\begin{table}[t]
  \centering
  \caption{Results of retriever pre-training. B is batch size, Q is the form of question used in training, HD is the number of hard negatives.}
    \vspace{-0.1in}
    \begin{tabular}{cccccc}
    \toprule
          & B & Q & HD & Recall@20 & Recall@100 \\
    \midrule
    BM25  & /     & /     & /     & 0.3711  & 0.5100  \\
    \midrule
    \multicolumn{1}{c}{\multirow{2}[2]{*}{ORConvQA }} & 16    & $q^{rw}$ & 0     & 0.1672  & 0.2916  \\
          & 16$\times$4  & $q^{rw}$ & 0     & 0.3561  & 0.5034  \\
    \midrule
    DPR   & 16    & $q^{rw}$ & 1     & 0.2395  & 0.3759  \\
    \midrule
    \multirow{2}[2]{*}{ours} & 16    & $q^{rw}$/$q^{rw}$ & 1     & 0.3214  & 0.4689  \\
          & 16    & $q^{rw}$/$q^{or}$ & 1     & 0.3690  & 0.5070  \\
          & 32    & $q^{rw}$/$q^{or}$ & 1     & 0.4675  & 0.5882  \\
    \bottomrule
    \end{tabular}%
    \vspace{-0.2in}
  \label{tab:resofpretraining}%
\end{table}%

\vspace{-0.1in}
\subsection{Further Analysis}
\subsubsection{ Retriever Performance}\label{subsec:fa1}
We evaluate some existing pre-trained retriever with our KL-divergence regularized retriever on the full collection of Wikipedia passages. The evaluation results are reported in Table~\ref{tab:resofpretraining}. 
As the typical sparse retriever, BM25 achieves a good retrieval performance while its QA performance is limited as it can not be fine-tuned in subsequent joint training.
Both ORConvQA and DPR retrievers take the question rewrite as input following the setting in~\cite{qu2020open}. The only difference between them is that DPR uses a TF-IDF hard negative provided by the dataset in addition to the in-batch negatives. DPR outperforms ORConvQA when the batch size is the same, which indicates that the hard negative play a key role during training. Considering the batch size of ORConvQA in~\cite{qu2020open} is 64, we also show the retrieval performance, which is improved dramatically as the number of in-batch negatives increases.
Our KL-divergence based regularized retriever uses two question forms, rewrite and original question, as input, and offers the best performance with the same batch size of 16 among these dense retrieval model. When the batch size is raised to 32, our retriever achieves a better performance and exceeds BM25. 
Moreover, to explore the advantage of the original question which is the concatenation of the historical and current questions, we evaluate our retriever by feeding two of the same rewrite. The results shows that employing two different forms of one question leads to a better performance. It is probably because the input of two same rewrites works just as a regularization while that of original question and rewrite is able to add a supervision for question understanding learning.  

\subsubsection{Case Study for Post-Ranker}
\begin{figure}[t]
    \centering
    \includegraphics[width=\linewidth]{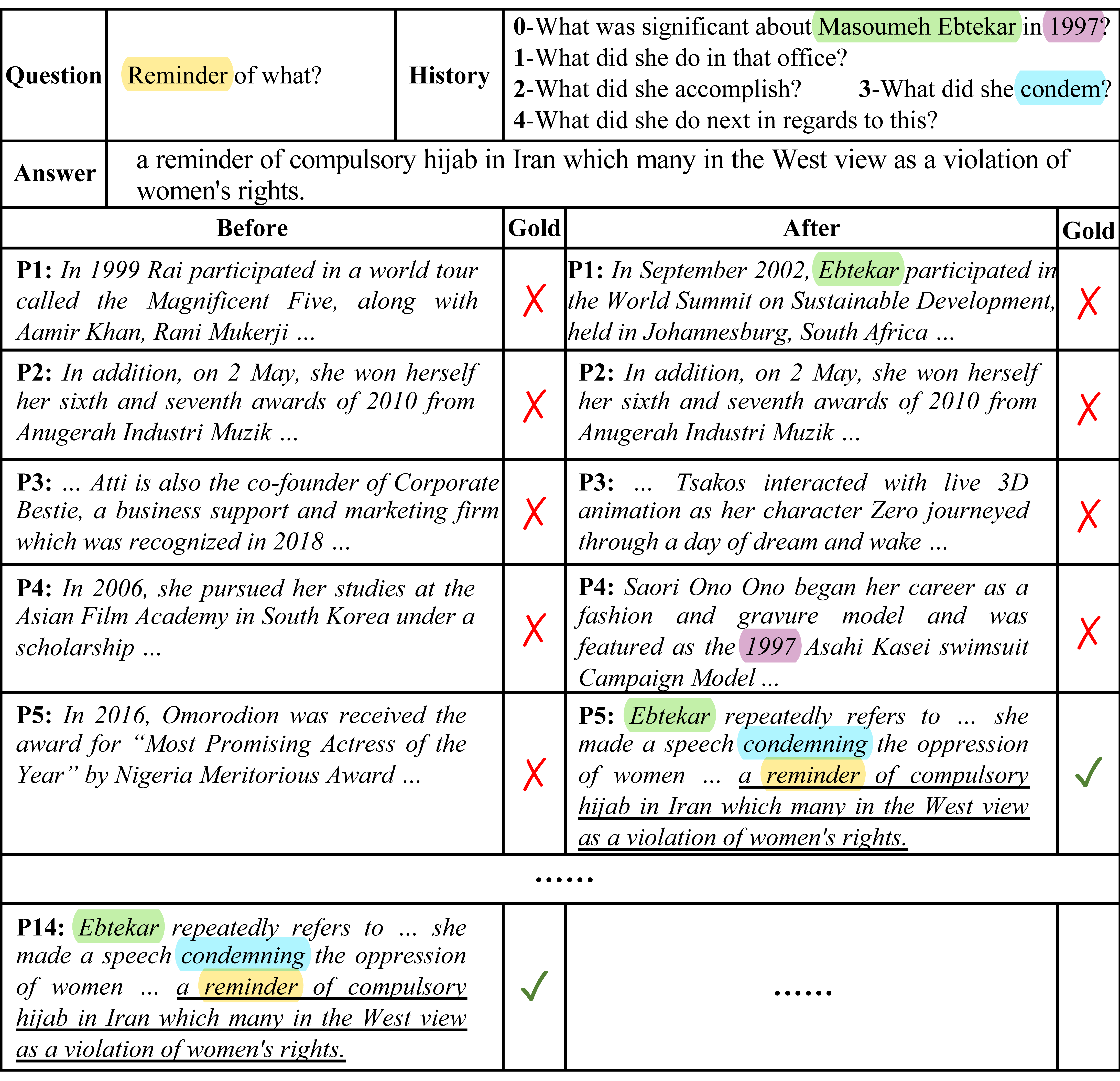}
    \vspace{-0.3in}
    \caption{Case study of the passage retrieval results before and after post-ranker for an example question.}
    \label{fig:case1}
    \vspace{-0.15in}
\end{figure}

To explore the effect of post-ranker, we count the number of golden passages before and after post-ranker in the inference phase. Concretely, the number of golden passages in the top $T$ ($T=5$) increases from 2,228 to 2,269 while the golden passages ranked between $T+1$ and $K$ ($K=100$) decreases from 1,292 to 1,251, which verifies post-ranker' ability in enforcing the relevant passages to appear in the higher place. Figure~\ref{fig:case1} shows an example of the passage retrieval results before and after post-ranker for a question. Before applying post-ranker, it can be observed that golden passage is not contained in the top 5 passages but ranked in the 14 place of the ranking list generated by the retriever. Taking the passage representations in the ranking list as input, post-ranker provides a new passage ranking where the golden passage is reranked in the top 5. Moreover, the top 5 passages of the newly generated ranking list contains more relevant content with the question and its historical questions, showing the advantage of the proposed post-ranker module in our {\modelname}.  


\subsubsection{Impact of Curriculum Learning Strategy}

\begin{figure}[t]
     \centering
     \begin{subfigure}[b]{0.23\textwidth}
         \centering
         \includegraphics[width=\textwidth]{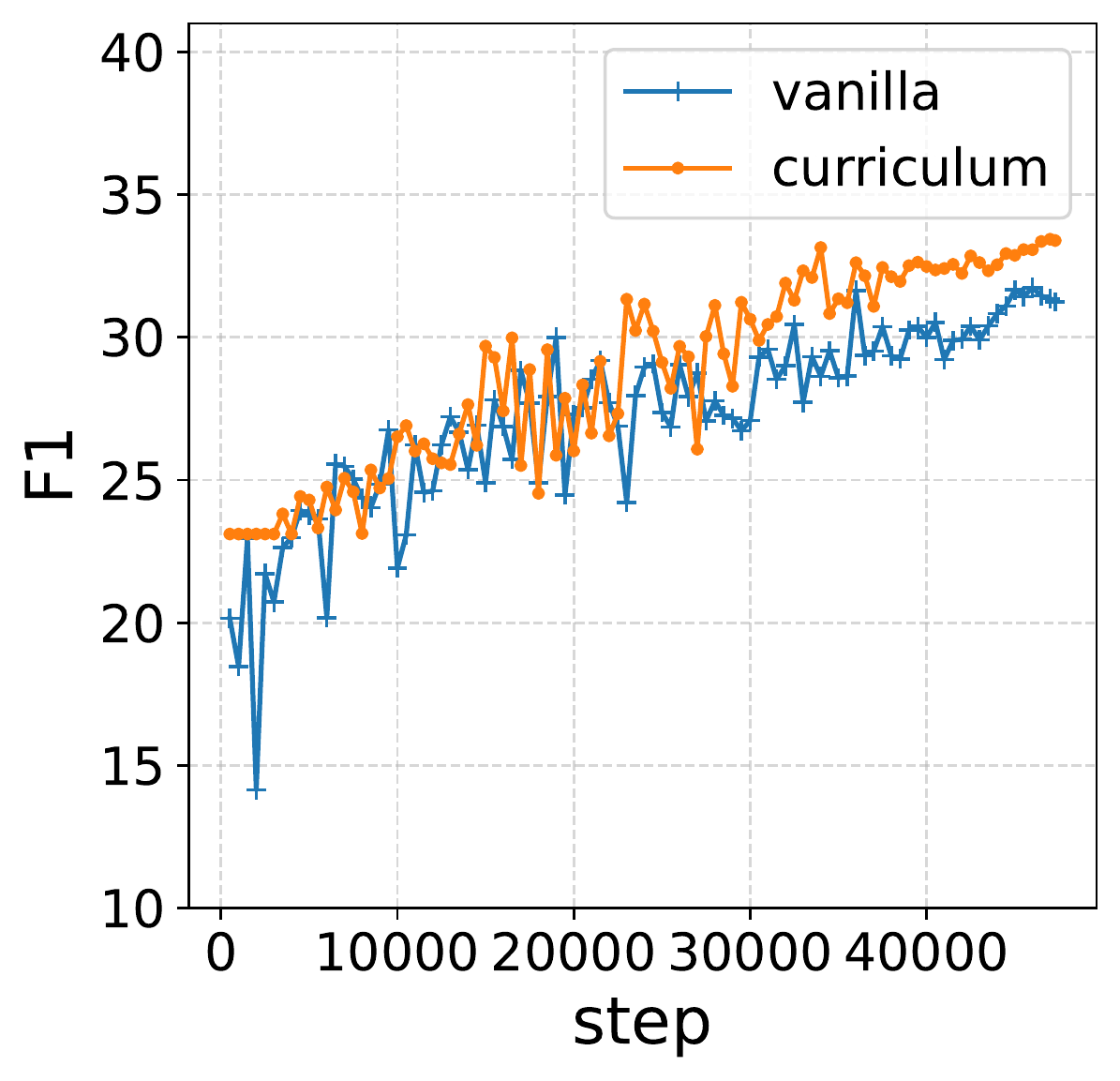}
         \caption{}
         \label{fig:curriculum_f1}
     \end{subfigure}
     \hfill
     \begin{subfigure}[b]{0.23\textwidth}
         \centering
         \includegraphics[width=\textwidth]{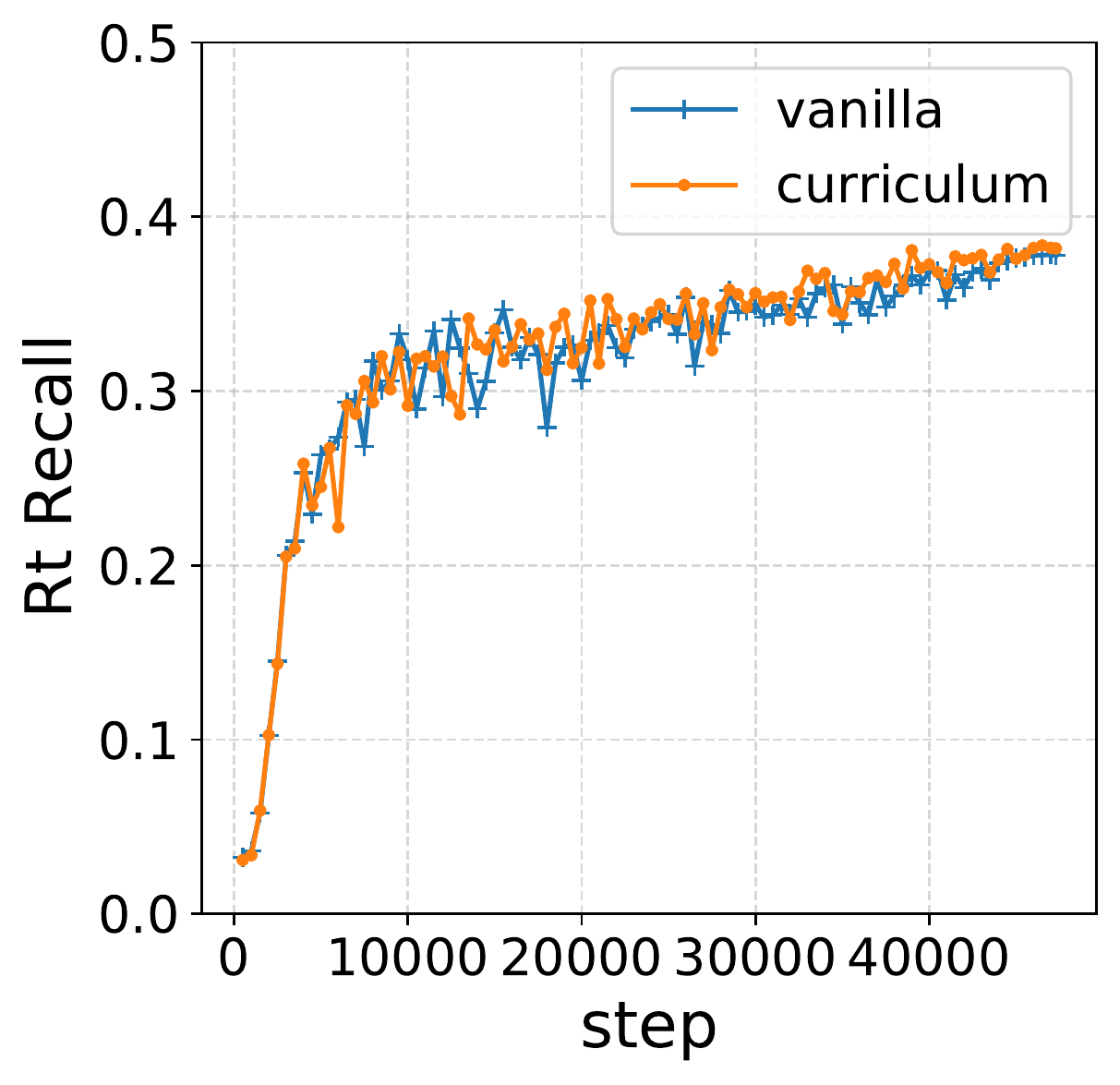}
         \caption{}
         \label{fig:curriculum_recall}
     \end{subfigure}
     \vspace{-0.15in}
        \caption{Comparison of performance improvement with / without curriculum learning.}
    \vspace{-0.2in}
        \label{fig:curriculum}
\end{figure}

To investigate the impact of curriculum learning, we save checkpoints and evaluate on the test set every 500 steps, the results of all the checkpoints are shown in Figure~\ref{fig:curriculum}.
It can be observed that the QA metric F1 with curriculum learning increases more steadily and reached better final performance faster. 
For retrieval metric, the improvement achieved by curriculum learning is the higher final recall value, which is not quite noticeable.
It indicates that the curriculum learning strategy makes more pronounced contribution for the answer reading other than the passage retrieval, which in reasonable as the inconsistent problem addressed by curriculum learning mainly reflects in the reading part.

\vspace{-0.1in}
\section{Related Work}\label{sec:related}

This section briefly summarizes some existing works on open domain question answering and conversational question answering, which are most relevant to this work.

\vspace{-0.1in}
\subsection{Open Domain Question Answering}
Open-domain question answering (OpenQA) \cite{voorhees1999trec} is a task that uses a huge library of documents to answer factual queries. The two-stage design, which included a passage retriever to choose a subset of passages and a machine reader to exact answers, became popular after DrQA~\cite{chen2017reading}.

The passage retriever is an important component of OpenQA system since it searches relevant paragraphs for the next stage. 
Traditional sparse retrieval models, such as TF-IDF or BM25~\cite{robertson2009probabilistic}, have been widely adopted as retriever in OpenQA systems~\cite{chen2017reading, yang2019end, lin2018denoising}. While sparse retrieval cannot handle the case of high semantic correlation with little lexical overlap and it is untrainable, dense passage retrievers have lately gained popularity~\cite{lee2019latent, guu2020realm, karpukhin2020dense, Qu2021RocketQAAO}. 
In general, the dense retrieval model is a dual-encoder architecture that encodes both the question and the passage individually. Both encoders are trained during the retriever pre-training process. When training with the reader for the QA task, only the question encoder is normally fine-tuned.
In order to increase the retrieval impact of dense retrievers, some studies incorporate hard negatives. 
BM25 top passages which do not contain answers are utilized as hard negatives~\cite{karpukhin2020dense, gao2020complementing}. And dynamic hard negatives are employed in~\cite{xiong2021approximate, zhan2021optimizing,guu2020realm}, which are the top-ranked irrelevant documents given by dense retriever during training. 

A contemporary OpenQA system also includes a reader as a key component. Its goal is to deduce the answer to a query from a collection of documents. Existing Readers may be divided into two types: (1) Extractive readers~\cite{chen2017reading,yang2019end,karpukhin2020dense}, which anticipate an answer span from the retrieved texts, (2) Generative readers~\cite{lewis2020retrieval,izacard2021leveraging}, which produce natural language replies using sequence-to-sequence (Seq2Seq) models.

\subsection{Conversational Question Answering} 
Conversational Question Answering (CQA) is required to understand the given context and history dialogue to answer the question.
As a main type of CQA, Conversational Machine Reading Comprehension (CMRC)~\cite{qu2019bert, qu2019attentive, qiu2021reinforced} does the QA task with text-based corpora. 
For CMRC, the number of conversational history turns is critical, as context utterances that are relevant to the inquiry are valuable, while irrelevant ones may introduce additional noise. For example, \cite{qu2019bert, qu2020open} makes use of conversation history by including $K$ rounds of history turns. \cite{qu2019attentive} weights previous conversation rounds based on their contribution to the answer to the current question.

The approaches outlined above rely extensively either on the provided material or a given paragraph to extract or generate answers. 
However, this is impractical in real world since golden passage is not always available. 
Open retrieval methods which try to obtain evidence from a big collection, have lately been popular in the CMRC. 
ORConvQA~\cite{qu2020open} is the first work proposing three primary modules for open-retrieval CQA: (1) a passage retriever, (2) a passage reranker, and (3) a passage reader. Given a query and its conversational history, the passage retriever retrieves the top $K$ relevant texts from a large-scale corpus. The passage reranker and reader then rerank and read the top texts to discover the correct answer. 
The research for the conversational OpenQA needs to be further explored. This work tries to improve ORConvQA from multiple aspects including regularizing retriever pre-training, incorporating post-ranking, and curriculum learning.

\section{Conclusion}\label{sec:conclusion}
This paper proposes multifaceted improvements for the conversational OpenQA task. Concretely, a KL-divergence based regularization is proposed in pre-training for a better question understanding. A post-ranker module is added to realize the joint training of question and passage representations to generate a better passage ranking. A semi-automatic curriculum learning strategy is designed to encourage the reader to find the answer without manually adding golden passages.
The experimental evaluation demonstrates the effectiveness of our {\modelname}.

\newpage
\bibliographystyle{ACM-Reference-Format}
\bibliography{sample-base}


\begin{thebibliography}{34}


\ifx \showCODEN    \undefined \def \showCODEN     #1{\unskip}     \fi
\ifx \showDOI      \undefined \def \showDOI       #1{#1}\fi
\ifx \showISBNx    \undefined \def \showISBNx     #1{\unskip}     \fi
\ifx \showISBNxiii \undefined \def \showISBNxiii  #1{\unskip}     \fi
\ifx \showISSN     \undefined \def \showISSN      #1{\unskip}     \fi
\ifx \showLCCN     \undefined \def \showLCCN      #1{\unskip}     \fi
\ifx \shownote     \undefined \def \shownote      #1{#1}          \fi
\ifx \showarticletitle \undefined \def \showarticletitle #1{#1}   \fi
\ifx \showURL      \undefined \def \showURL       {\relax}        \fi
\providecommand\bibfield[2]{#2}
\providecommand\bibinfo[2]{#2}
\providecommand\natexlab[1]{#1}
\providecommand\showeprint[2][]{arXiv:#2}

\bibitem[\protect\citeauthoryear{Bengio, Louradour, Collobert, and
  Weston}{Bengio et~al\mbox{.}}{2009}]%
        {bengio2009curriculum}
\bibfield{author}{\bibinfo{person}{Yoshua Bengio},
  \bibinfo{person}{J{\'e}r{\^o}me Louradour}, \bibinfo{person}{Ronan
  Collobert}, {and} \bibinfo{person}{Jason Weston}.}
  \bibinfo{year}{2009}\natexlab{}.
\newblock \showarticletitle{Curriculum learning}. In
  \bibinfo{booktitle}{\emph{Proceedings of the 26th Annual International
  Conference on Machine Learning}}. \bibinfo{pages}{41--48}.
\newblock


\bibitem[\protect\citeauthoryear{Bromley, Bentz, Bottou, Guyon, LeCun, Moore,
  S{\"a}ckinger, and Shah}{Bromley et~al\mbox{.}}{1993}]%
        {bromley1993signature}
\bibfield{author}{\bibinfo{person}{Jane Bromley}, \bibinfo{person}{James~W
  Bentz}, \bibinfo{person}{L{\'e}on Bottou}, \bibinfo{person}{Isabelle Guyon},
  \bibinfo{person}{Yann LeCun}, \bibinfo{person}{Cliff Moore},
  \bibinfo{person}{Eduard S{\"a}ckinger}, {and} \bibinfo{person}{Roopak Shah}.}
  \bibinfo{year}{1993}\natexlab{}.
\newblock \showarticletitle{Signature verification using a “siamese” time
  delay neural network}.
\newblock \bibinfo{journal}{\emph{International Journal of Pattern Recognition
  and Artificial Intelligence}} \bibinfo{volume}{7}, \bibinfo{number}{04}
  (\bibinfo{year}{1993}), \bibinfo{pages}{669--688}.
\newblock


\bibitem[\protect\citeauthoryear{Chen, Fisch, Weston, and Bordes}{Chen
  et~al\mbox{.}}{2017}]%
        {chen2017reading}
\bibfield{author}{\bibinfo{person}{Danqi Chen}, \bibinfo{person}{Adam Fisch},
  \bibinfo{person}{Jason Weston}, {and} \bibinfo{person}{Antoine Bordes}.}
  \bibinfo{year}{2017}\natexlab{}.
\newblock \showarticletitle{Reading Wikipedia to Answer Open-Domain Questions}.
  In \bibinfo{booktitle}{\emph{Proceedings of the 55th Annual Meeting of the
  Association for Computational Linguistics}}. \bibinfo{pages}{1870--1879}.
\newblock


\bibitem[\protect\citeauthoryear{Choi, He, Iyyer, Yatskar, Yih, Choi, Liang,
  and Zettlemoyer}{Choi et~al\mbox{.}}{2018}]%
        {choi2018quac}
\bibfield{author}{\bibinfo{person}{Eunsol Choi}, \bibinfo{person}{He He},
  \bibinfo{person}{Mohit Iyyer}, \bibinfo{person}{Mark Yatskar},
  \bibinfo{person}{Wen-tau Yih}, \bibinfo{person}{Yejin Choi},
  \bibinfo{person}{Percy Liang}, {and} \bibinfo{person}{Luke Zettlemoyer}.}
  \bibinfo{year}{2018}\natexlab{}.
\newblock \showarticletitle{QuAC: Question Answering in Context}. In
  \bibinfo{booktitle}{\emph{Proceedings of the 2018 Conference on Empirical
  Methods in Natural Language Processing}}. \bibinfo{pages}{2174--2184}.
\newblock


\bibitem[\protect\citeauthoryear{Devlin, Chang, Lee, and Toutanova}{Devlin
  et~al\mbox{.}}{2019}]%
        {devlin2018bert}
\bibfield{author}{\bibinfo{person}{Jacob Devlin}, \bibinfo{person}{Ming-Wei
  Chang}, \bibinfo{person}{Kenton Lee}, {and} \bibinfo{person}{Kristina
  Toutanova}.} \bibinfo{year}{2019}\natexlab{}.
\newblock \showarticletitle{BERT: Pre-training of Deep Bidirectional
  Transformers for Language Understanding}.
\newblock  (\bibinfo{year}{2019}), \bibinfo{pages}{4171--4186}.
\newblock


\bibitem[\protect\citeauthoryear{Elgohary, Peskov, and Boyd-Graber}{Elgohary
  et~al\mbox{.}}{2019}]%
        {elgohary2019can}
\bibfield{author}{\bibinfo{person}{Ahmed Elgohary}, \bibinfo{person}{Denis
  Peskov}, {and} \bibinfo{person}{Jordan Boyd-Graber}.}
  \bibinfo{year}{2019}\natexlab{}.
\newblock \showarticletitle{Can You Unpack That? Learning to Rewrite
  Questions-in-Context}. In \bibinfo{booktitle}{\emph{Proceedings of the 2019
  Conference on Empirical Methods in Natural Language Processing and the 9th
  International Joint Conference on Natural Language Processing
  (EMNLP-IJCNLP)}}. \bibinfo{pages}{5918--5924}.
\newblock


\bibitem[\protect\citeauthoryear{Gao, Dai, Chen, Fan, Van~Durme, and
  Callan}{Gao et~al\mbox{.}}{2020}]%
        {gao2020complementing}
\bibfield{author}{\bibinfo{person}{Luyu Gao}, \bibinfo{person}{Zhuyun Dai},
  \bibinfo{person}{Tongfei Chen}, \bibinfo{person}{Zhen Fan},
  \bibinfo{person}{Benjamin Van~Durme}, {and} \bibinfo{person}{Jamie Callan}.}
  \bibinfo{year}{2020}\natexlab{}.
\newblock \showarticletitle{Complementing lexical retrieval with semantic
  residual embedding}.
\newblock \bibinfo{journal}{\emph{arXiv preprint arXiv:2004.13969}}
  (\bibinfo{year}{2020}).
\newblock


\bibitem[\protect\citeauthoryear{Guu, Lee, Tung, Pasupat, and Chang}{Guu
  et~al\mbox{.}}{2020}]%
        {guu2020realm}
\bibfield{author}{\bibinfo{person}{Kelvin Guu}, \bibinfo{person}{Kenton Lee},
  \bibinfo{person}{Zora Tung}, \bibinfo{person}{Panupong Pasupat}, {and}
  \bibinfo{person}{Ming-Wei Chang}.} \bibinfo{year}{2020}\natexlab{}.
\newblock \showarticletitle{Realm: Retrieval-augmented language model
  pre-training}.
\newblock \bibinfo{journal}{\emph{arXiv preprint arXiv:2002.08909}}
  (\bibinfo{year}{2020}).
\newblock


\bibitem[\protect\citeauthoryear{Izacard and Grave}{Izacard and Grave}{2021}]%
        {izacard2021leveraging}
\bibfield{author}{\bibinfo{person}{Gautier Izacard} {and}
  \bibinfo{person}{{\'E}douard Grave}.} \bibinfo{year}{2021}\natexlab{}.
\newblock \showarticletitle{Leveraging Passage Retrieval with Generative Models
  for Open Domain Question Answering}. In \bibinfo{booktitle}{\emph{Proceedings
  of the 16th Conference of the European Chapter of the Association for
  Computational Linguistics: Main Volume}}. \bibinfo{pages}{874--880}.
\newblock


\bibitem[\protect\citeauthoryear{Johnson, Douze, and J{\'e}gou}{Johnson
  et~al\mbox{.}}{2017}]%
        {JDH17}
\bibfield{author}{\bibinfo{person}{Jeff Johnson}, \bibinfo{person}{Matthijs
  Douze}, {and} \bibinfo{person}{Herv{\'e} J{\'e}gou}.}
  \bibinfo{year}{2017}\natexlab{}.
\newblock \showarticletitle{Billion-scale similarity search with GPUs}.
\newblock \bibinfo{journal}{\emph{arXiv preprint arXiv:1702.08734}}
  (\bibinfo{year}{2017}).
\newblock


\bibitem[\protect\citeauthoryear{Karpukhin, Oguz, Min, Lewis, Wu, Edunov, Chen,
  and Yih}{Karpukhin et~al\mbox{.}}{2020}]%
        {karpukhin2020dense}
\bibfield{author}{\bibinfo{person}{Vladimir Karpukhin}, \bibinfo{person}{Barlas
  Oguz}, \bibinfo{person}{Sewon Min}, \bibinfo{person}{Patrick Lewis},
  \bibinfo{person}{Ledell Wu}, \bibinfo{person}{Sergey Edunov},
  \bibinfo{person}{Danqi Chen}, {and} \bibinfo{person}{Wen-tau Yih}.}
  \bibinfo{year}{2020}\natexlab{}.
\newblock \showarticletitle{Dense Passage Retrieval for Open-Domain Question
  Answering}. In \bibinfo{booktitle}{\emph{Proceedings of the 2020 Conference
  on Empirical Methods in Natural Language Processing (EMNLP)}}.
  \bibinfo{pages}{6769--6781}.
\newblock


\bibitem[\protect\citeauthoryear{Kenton and Toutanova}{Kenton and
  Toutanova}{2019}]%
        {kenton2019bert}
\bibfield{author}{\bibinfo{person}{Jacob Devlin Ming-Wei~Chang Kenton} {and}
  \bibinfo{person}{Lee~Kristina Toutanova}.} \bibinfo{year}{2019}\natexlab{}.
\newblock \showarticletitle{Bert: Pre-training of deep bidirectional
  transformers for language understanding}. In
  \bibinfo{booktitle}{\emph{Proceedings of NAACL-HLT}}.
  \bibinfo{pages}{4171--4186}.
\newblock


\bibitem[\protect\citeauthoryear{Kratzwald and Feuerriegel}{Kratzwald and
  Feuerriegel}{2018}]%
        {kratzwald2018adaptive}
\bibfield{author}{\bibinfo{person}{Bernhard Kratzwald} {and}
  \bibinfo{person}{Stefan Feuerriegel}.} \bibinfo{year}{2018}\natexlab{}.
\newblock \showarticletitle{Adaptive Document Retrieval for Deep Question
  Answering}. In \bibinfo{booktitle}{\emph{Proceedings of the 2018 Conference
  on Empirical Methods in Natural Language Processing}}.
  \bibinfo{pages}{576--581}.
\newblock


\bibitem[\protect\citeauthoryear{Lan, Chen, Goodman, Gimpel, Sharma, and
  Soricut}{Lan et~al\mbox{.}}{2020}]%
        {lan2020albert}
\bibfield{author}{\bibinfo{person}{Zhenzhong Lan}, \bibinfo{person}{Mingda
  Chen}, \bibinfo{person}{Sebastian Goodman}, \bibinfo{person}{Kevin Gimpel},
  \bibinfo{person}{Piyush Sharma}, {and} \bibinfo{person}{Radu Soricut}.}
  \bibinfo{year}{2020}\natexlab{}.
\newblock \bibinfo{title}{ALBERT: A Lite BERT for Self-supervised Learning of
  Language Representations}.
\newblock
\newblock
\showeprint[arxiv]{1909.11942}~[cs.CL]


\bibitem[\protect\citeauthoryear{Lee, Chang, and Toutanova}{Lee
  et~al\mbox{.}}{2019}]%
        {lee2019latent}
\bibfield{author}{\bibinfo{person}{Kenton Lee}, \bibinfo{person}{Ming-Wei
  Chang}, {and} \bibinfo{person}{Kristina Toutanova}.}
  \bibinfo{year}{2019}\natexlab{}.
\newblock \showarticletitle{Latent retrieval for weakly supervised open domain
  question answering}.
\newblock \bibinfo{journal}{\emph{arXiv preprint arXiv:1906.00300}}
  (\bibinfo{year}{2019}).
\newblock


\bibitem[\protect\citeauthoryear{Lewis, Perez, Piktus, Petroni, Karpukhin,
  Goyal, K{\"u}ttler, Lewis, Yih, Rockt{\"a}schel, et~al\mbox{.}}{Lewis
  et~al\mbox{.}}{2020}]%
        {lewis2020retrieval}
\bibfield{author}{\bibinfo{person}{Patrick Lewis}, \bibinfo{person}{Ethan
  Perez}, \bibinfo{person}{Aleksandra Piktus}, \bibinfo{person}{Fabio Petroni},
  \bibinfo{person}{Vladimir Karpukhin}, \bibinfo{person}{Naman Goyal},
  \bibinfo{person}{Heinrich K{\"u}ttler}, \bibinfo{person}{Mike Lewis},
  \bibinfo{person}{Wen-tau Yih}, \bibinfo{person}{Tim Rockt{\"a}schel},
  {et~al\mbox{.}}} \bibinfo{year}{2020}\natexlab{}.
\newblock \showarticletitle{Retrieval-augmented generation for
  knowledge-intensive nlp tasks}.
\newblock \bibinfo{journal}{\emph{arXiv preprint arXiv:2005.11401}}
  (\bibinfo{year}{2020}).
\newblock


\bibitem[\protect\citeauthoryear{Lin, Ji, Liu, and Sun}{Lin
  et~al\mbox{.}}{2018}]%
        {lin2018denoising}
\bibfield{author}{\bibinfo{person}{Yankai Lin}, \bibinfo{person}{Haozhe Ji},
  \bibinfo{person}{Zhiyuan Liu}, {and} \bibinfo{person}{Maosong Sun}.}
  \bibinfo{year}{2018}\natexlab{}.
\newblock \showarticletitle{Denoising distantly supervised open-domain question
  answering}. In \bibinfo{booktitle}{\emph{Proceedings of the 56th Annual
  Meeting of the Association for Computational Linguistics (Volume 1: Long
  Papers)}}. \bibinfo{pages}{1736--1745}.
\newblock


\bibitem[\protect\citeauthoryear{Liu, Ott, Goyal, Du, Joshi, Chen, Levy, Lewis,
  Zettlemoyer, and Stoyanov}{Liu et~al\mbox{.}}{2019}]%
        {liu2019roberta}
\bibfield{author}{\bibinfo{person}{Yinhan Liu}, \bibinfo{person}{Myle Ott},
  \bibinfo{person}{Naman Goyal}, \bibinfo{person}{Jingfei Du},
  \bibinfo{person}{Mandar Joshi}, \bibinfo{person}{Danqi Chen},
  \bibinfo{person}{Omer Levy}, \bibinfo{person}{Mike Lewis},
  \bibinfo{person}{Luke Zettlemoyer}, {and} \bibinfo{person}{Veselin
  Stoyanov}.} \bibinfo{year}{2019}\natexlab{}.
\newblock \showarticletitle{Roberta: A robustly optimized bert pretraining
  approach}.
\newblock \bibinfo{journal}{\emph{arXiv preprint arXiv:1907.11692}}
  (\bibinfo{year}{2019}).
\newblock


\bibitem[\protect\citeauthoryear{Min, Chen, Zettlemoyer, and Hajishirzi}{Min
  et~al\mbox{.}}{2019}]%
        {min2019knowledge}
\bibfield{author}{\bibinfo{person}{Sewon Min}, \bibinfo{person}{Danqi Chen},
  \bibinfo{person}{Luke Zettlemoyer}, {and} \bibinfo{person}{Hannaneh
  Hajishirzi}.} \bibinfo{year}{2019}\natexlab{}.
\newblock \showarticletitle{Knowledge guided text retrieval and reading for
  open domain question answering}.
\newblock \bibinfo{journal}{\emph{arXiv preprint arXiv:1911.03868}}
  (\bibinfo{year}{2019}).
\newblock


\bibitem[\protect\citeauthoryear{Qiu, Huang, Chen, Ji, Qu, Wei, Huang, and
  Zhang}{Qiu et~al\mbox{.}}{2021}]%
        {qiu2021reinforced}
\bibfield{author}{\bibinfo{person}{Minghui Qiu}, \bibinfo{person}{Xinjing
  Huang}, \bibinfo{person}{Cen Chen}, \bibinfo{person}{Feng Ji},
  \bibinfo{person}{Chen Qu}, \bibinfo{person}{Wei Wei}, \bibinfo{person}{Jun
  Huang}, {and} \bibinfo{person}{Yin Zhang}.} \bibinfo{year}{2021}\natexlab{}.
\newblock \showarticletitle{Reinforced history backtracking for conversational
  question answering}. In \bibinfo{booktitle}{\emph{Proceedings of the 35th
  Conference on Artificial Intelligence, AAAI}}.
\newblock


\bibitem[\protect\citeauthoryear{Qu, Yang, Chen, Qiu, Croft, and Iyyer}{Qu
  et~al\mbox{.}}{2020}]%
        {qu2020open}
\bibfield{author}{\bibinfo{person}{Chen Qu}, \bibinfo{person}{Liu Yang},
  \bibinfo{person}{Cen Chen}, \bibinfo{person}{Minghui Qiu},
  \bibinfo{person}{W~Bruce Croft}, {and} \bibinfo{person}{Mohit Iyyer}.}
  \bibinfo{year}{2020}\natexlab{}.
\newblock \showarticletitle{Open-retrieval conversational question answering}.
  In \bibinfo{booktitle}{\emph{Proceedings of the 43rd International ACM SIGIR
  Conference on Research and Development in Information Retrieval}}.
  \bibinfo{pages}{539--548}.
\newblock


\bibitem[\protect\citeauthoryear{Qu, Yang, Qiu, Croft, Zhang, and Iyyer}{Qu
  et~al\mbox{.}}{2019a}]%
        {qu2019bert}
\bibfield{author}{\bibinfo{person}{Chen Qu}, \bibinfo{person}{Liu Yang},
  \bibinfo{person}{Minghui Qiu}, \bibinfo{person}{W~Bruce Croft},
  \bibinfo{person}{Yongfeng Zhang}, {and} \bibinfo{person}{Mohit Iyyer}.}
  \bibinfo{year}{2019}\natexlab{a}.
\newblock \showarticletitle{BERT with history answer embedding for
  conversational question answering}. In \bibinfo{booktitle}{\emph{Proceedings
  of the 42nd international ACM SIGIR conference on research and development in
  information retrieval}}. \bibinfo{pages}{1133--1136}.
\newblock


\bibitem[\protect\citeauthoryear{Qu, Yang, Qiu, Zhang, Chen, Croft, and
  Iyyer}{Qu et~al\mbox{.}}{2019b}]%
        {qu2019attentive}
\bibfield{author}{\bibinfo{person}{Chen Qu}, \bibinfo{person}{Liu Yang},
  \bibinfo{person}{Minghui Qiu}, \bibinfo{person}{Yongfeng Zhang},
  \bibinfo{person}{Cen Chen}, \bibinfo{person}{W~Bruce Croft}, {and}
  \bibinfo{person}{Mohit Iyyer}.} \bibinfo{year}{2019}\natexlab{b}.
\newblock \showarticletitle{Attentive history selection for conversational
  question answering}. In \bibinfo{booktitle}{\emph{Proceedings of the 28th ACM
  International Conference on Information and Knowledge Management}}.
  \bibinfo{pages}{1391--1400}.
\newblock


\bibitem[\protect\citeauthoryear{Qu, Ding, Liu, Liu, Ren, Zhao, Dong, Wu, and
  Wang}{Qu et~al\mbox{.}}{2021}]%
        {Qu2021RocketQAAO}
\bibfield{author}{\bibinfo{person}{Yingqi Qu}, \bibinfo{person}{Yuchen Ding},
  \bibinfo{person}{Jing Liu}, \bibinfo{person}{Kai Liu},
  \bibinfo{person}{Ruiyang Ren}, \bibinfo{person}{Wayne~Xin Zhao},
  \bibinfo{person}{Daxiang Dong}, \bibinfo{person}{Hua Wu}, {and}
  \bibinfo{person}{Haifeng Wang}.} \bibinfo{year}{2021}\natexlab{}.
\newblock \showarticletitle{RocketQA: An optimized training approach to dense
  passage retrieval for open-domain question answering}. In
  \bibinfo{booktitle}{\emph{Proceedings of the 2021 Conference of the North
  American Chapter of the Association for Computational Linguistics: Human
  Language Technologies}}. \bibinfo{pages}{5835--5847}.
\newblock


\bibitem[\protect\citeauthoryear{Robertson and Zaragoza}{Robertson and
  Zaragoza}{2009}]%
        {robertson2009probabilistic}
\bibfield{author}{\bibinfo{person}{Stephen Robertson} {and}
  \bibinfo{person}{Hugo Zaragoza}.} \bibinfo{year}{2009}\natexlab{}.
\newblock \bibinfo{booktitle}{\emph{The probabilistic relevance framework: BM25
  and beyond}}.
\newblock \bibinfo{publisher}{Now Publishers Inc}.
\newblock


\bibitem[\protect\citeauthoryear{Santos, Tan, Xiang, and Zhou}{Santos
  et~al\mbox{.}}{2016}]%
        {santos2016attentive}
\bibfield{author}{\bibinfo{person}{Cicero~dos Santos}, \bibinfo{person}{Ming
  Tan}, \bibinfo{person}{Bing Xiang}, {and} \bibinfo{person}{Bowen Zhou}.}
  \bibinfo{year}{2016}\natexlab{}.
\newblock \showarticletitle{Attentive pooling networks}.
\newblock \bibinfo{journal}{\emph{arXiv preprint arXiv:1602.03609}}
  (\bibinfo{year}{2016}).
\newblock


\bibitem[\protect\citeauthoryear{Spitkovsky, Alshawi, and Jurafsky}{Spitkovsky
  et~al\mbox{.}}{2010}]%
        {spitkovsky2010baby}
\bibfield{author}{\bibinfo{person}{Valentin~I Spitkovsky},
  \bibinfo{person}{Hiyan Alshawi}, {and} \bibinfo{person}{Dan Jurafsky}.}
  \bibinfo{year}{2010}\natexlab{}.
\newblock \showarticletitle{From baby steps to leapfrog: How “less is more”
  in unsupervised dependency parsing}. In \bibinfo{booktitle}{\emph{Human
  Language Technologies: The 2010 Annual Conference of the North American
  Chapter of the Association for Computational Linguistics}}.
  \bibinfo{pages}{751--759}.
\newblock


\bibitem[\protect\citeauthoryear{Voorhees et~al\mbox{.}}{Voorhees
  et~al\mbox{.}}{1999}]%
        {voorhees1999trec}
\bibfield{author}{\bibinfo{person}{Ellen~M Voorhees} {et~al\mbox{.}}}
  \bibinfo{year}{1999}\natexlab{}.
\newblock \showarticletitle{The TREC-8 question answering track report}. In
  \bibinfo{booktitle}{\emph{Trec}}, Vol.~\bibinfo{volume}{99}. Citeseer,
  \bibinfo{pages}{77--82}.
\newblock


\bibitem[\protect\citeauthoryear{Weinberger, Blitzer, and Saul}{Weinberger
  et~al\mbox{.}}{2006}]%
        {weinberger2006distance}
\bibfield{author}{\bibinfo{person}{Kilian~Q Weinberger}, \bibinfo{person}{John
  Blitzer}, {and} \bibinfo{person}{Lawrence~K Saul}.}
  \bibinfo{year}{2006}\natexlab{}.
\newblock \showarticletitle{Distance metric learning for large margin nearest
  neighbor classification}. In \bibinfo{booktitle}{\emph{Advances in neural
  information processing systems}}. \bibinfo{pages}{1473--1480}.
\newblock


\bibitem[\protect\citeauthoryear{Xiong, Xiong, Li, Tang, Liu, Bennett, Ahmed,
  and Overwijk}{Xiong et~al\mbox{.}}{2021}]%
        {xiong2021approximate}
\bibfield{author}{\bibinfo{person}{Lee Xiong}, \bibinfo{person}{Chenyan Xiong},
  \bibinfo{person}{Ye Li}, \bibinfo{person}{Kwok-Fung Tang},
  \bibinfo{person}{Jialin Liu}, \bibinfo{person}{Paul Bennett},
  \bibinfo{person}{Junaid Ahmed}, {and} \bibinfo{person}{Arnold Overwijk}.}
  \bibinfo{year}{2021}\natexlab{}.
\newblock \showarticletitle{Approximate Nearest Neighbor Negative Contrastive
  Learning for Dense Text Retrieval}. In
  \bibinfo{booktitle}{\emph{International Conference on Learning
  Representations (ICLR)}}.
\newblock
\urldef\tempurl%
\url{https://www.microsoft.com/en-us/research/publication/approximate-nearest-neighbor-negative-contrastive-learning-for-dense-text-retrieval/}
\showURL{%
\tempurl}


\bibitem[\protect\citeauthoryear{Xiong, Li, Iyer, Du, Lewis, Wang, Mehdad, Yih,
  Riedel, Kiela, et~al\mbox{.}}{Xiong et~al\mbox{.}}{2020}]%
        {xiong2020answering}
\bibfield{author}{\bibinfo{person}{Wenhan Xiong}, \bibinfo{person}{Xiang Li},
  \bibinfo{person}{Srini Iyer}, \bibinfo{person}{Jingfei Du},
  \bibinfo{person}{Patrick Lewis}, \bibinfo{person}{William~Yang Wang},
  \bibinfo{person}{Yashar Mehdad}, \bibinfo{person}{Scott Yih},
  \bibinfo{person}{Sebastian Riedel}, \bibinfo{person}{Douwe Kiela},
  {et~al\mbox{.}}} \bibinfo{year}{2020}\natexlab{}.
\newblock \showarticletitle{Answering Complex Open-Domain Questions with
  Multi-Hop Dense Retrieval}. In \bibinfo{booktitle}{\emph{International
  Conference on Learning Representations}}.
\newblock


\bibitem[\protect\citeauthoryear{Yang, Xie, Lin, Li, Tan, Xiong, Li, and
  Lin}{Yang et~al\mbox{.}}{2019}]%
        {yang2019end}
\bibfield{author}{\bibinfo{person}{Wei Yang}, \bibinfo{person}{Yuqing Xie},
  \bibinfo{person}{Aileen Lin}, \bibinfo{person}{Xingyu Li},
  \bibinfo{person}{Luchen Tan}, \bibinfo{person}{Kun Xiong},
  \bibinfo{person}{Ming Li}, {and} \bibinfo{person}{Jimmy Lin}.}
  \bibinfo{year}{2019}\natexlab{}.
\newblock \showarticletitle{End-to-End Open-Domain Question Answering with
  BERTserini}. In \bibinfo{booktitle}{\emph{Proceedings of the 2019 Conference
  of the North American Chapter of the Association for Computational
  Linguistics}}. \bibinfo{pages}{72--77}.
\newblock


\bibitem[\protect\citeauthoryear{Zhan, Mao, Liu, Guo, Zhang, and Ma}{Zhan
  et~al\mbox{.}}{2021}]%
        {zhan2021optimizing}
\bibfield{author}{\bibinfo{person}{Jingtao Zhan}, \bibinfo{person}{Jiaxin Mao},
  \bibinfo{person}{Yiqun Liu}, \bibinfo{person}{Jiafeng Guo},
  \bibinfo{person}{Min Zhang}, {and} \bibinfo{person}{Shaoping Ma}.}
  \bibinfo{year}{2021}\natexlab{}.
\newblock \showarticletitle{Optimizing Dense Retrieval Model Training with Hard
  Negatives}.
\newblock \bibinfo{journal}{\emph{arXiv preprint arXiv:2104.08051}}
  (\bibinfo{year}{2021}).
\newblock


\bibitem[\protect\citeauthoryear{Zhu, Lei, Wang, Zheng, Poria, and Chua}{Zhu
  et~al\mbox{.}}{2021}]%
        {zhu2021retrieving}
\bibfield{author}{\bibinfo{person}{Fengbin Zhu}, \bibinfo{person}{Wenqiang
  Lei}, \bibinfo{person}{Chao Wang}, \bibinfo{person}{Jianming Zheng},
  \bibinfo{person}{Soujanya Poria}, {and} \bibinfo{person}{Tat-Seng Chua}.}
  \bibinfo{year}{2021}\natexlab{}.
\newblock \showarticletitle{Retrieving and reading: A comprehensive survey on
  open-domain question answering}.
\newblock \bibinfo{journal}{\emph{arXiv preprint arXiv:2101.00774}}
  (\bibinfo{year}{2021}).
\newblock


\end{thebibliography}


\end{document}